\documentclass[11pt]{article}

\usepackage[final]{acl}

\usepackage{times}
\usepackage{latexsym}

\usepackage[T1]{fontenc}

\usepackage[utf8]{inputenc}

\usepackage{microtype}

\usepackage{inconsolata}

\usepackage{graphicx}

\usepackage{amsmath}
\usepackage{diagbox}
\usepackage{threeparttable}
\usepackage{multirow}
\usepackage{graphicx}

\usepackage[most]{tcolorbox}
\usepackage{enumitem}
\usepackage{xcolor}

\usepackage{graphicx}
\usepackage{subcaption}
\usepackage{enumitem}

\usepackage[utf8]{inputenc} 
\usepackage[T1]{fontenc}    
\usepackage{hyperref}       
\usepackage{url}            
\usepackage{booktabs}       
\usepackage{amsfonts}       
\usepackage{nicefrac}       
\usepackage{microtype}      
\usepackage{xcolor}         
\usepackage{arydshln}

\usepackage{booktabs,multirow,makecell,array,colortbl,xcolor}

\usepackage{tcolorbox}
\tcbuselibrary{breakable,skins}

\newcolumntype{L}[1]{>{\raggedright\arraybackslash}p{#1}}
\newcolumntype{C}[1]{>{\centering\arraybackslash}p{#1}}

\newcommand{\met}[1]{\textsf{#1}}
\newcommand{\bestmet}[1]{\textsf{\textbf{#1}}}

\usepackage[most]{tcolorbox}
\usepackage{xcolor}
\usepackage{enumitem}
\tcbuselibrary{breakable,skins}

\newtcolorbox{vlqa}{
  enhanced,
  breakable,
  colback=gray!3,
  colframe=black!50,
  boxrule=0.6pt,
  arc=3pt,
  left=6pt,right=6pt,top=4pt,bottom=4pt,
  fonttitle=\bfseries,
  before upper=\ttfamily\footnotesize, 
  title={VLM Predictions without Visual Input},
  sidebyside,
  sidebyside gap=6pt,
  sidebyside align=top seam,
  righthand width=0.36\linewidth,
}

\newtcolorbox{predcell}{
  enhanced,
  colback=white,
  colframe=black!45,
  boxrule=0.45pt,
  arc=2pt,
  boxsep=0.5pt,
  left=3pt,right=3pt,top=0.6pt,bottom=0.6pt,
}

\newcommand{\predcellrow}[2]{%
  \begin{predcell}\tiny\ttfamily #1\hfill \textbf{#2}\end{predcell}\vspace{-1.5pt}%
}
\newcommand{\predcellhuman}[2]{%
  \begin{predcell}\tiny\ttfamily #1\hfill #2\end{predcell}%
}

\usepackage{tcolorbox}
\tcbuselibrary{breakable}
\usepackage{fvextra}
\usepackage{xcolor} 

\DefineVerbatimEnvironment{PromptVerb}{Verbatim}{
  fontsize=\scriptsize,
  fontfamily=tt,
  breaklines=true,
  breakanywhere=true,
  breakautoindent=false,
  breaksymbolleft=\textcolor{gray}{\tiny\ensuremath{\hookrightarrow}}
}

\title{Are Vision Language Models Cross-Cultural Theory of Mind Reasoners?}

\author{
  Zabir Al Nazi$^{1}$ \quad
  GM~Shahariar$^{1}$ \quad
  Md. Abrar Hossain$^{2}$ \quad
  Wei Peng$^{3}$ \\
  $^{1}$University of California, Riverside \\
  $^{2}$University of Dhaka \\
  $^{3}$Stanford University \\
}

\begin{document}
\maketitle

\begin{abstract}
Theory of Mind (ToM) - the ability to attribute beliefs and intents to others - is fundamental for social intelligence, yet Vision-Language Model (VLM) evaluations remain largely Western-centric. In this work, we introduce \textbf{CulturalToM-VQA}, a benchmark of 5{,}095 visually situated ToM probes
across diverse cultural contexts, rituals, and social norms. Constructed through a frontier proprietary MLLM,
human-verified pipeline, the dataset spans a taxonomy of six ToM tasks and four complexity levels. We benchmark 10 VLMs (2023-2025) and observe a significant performance leap: while earlier models struggle, frontier models achieve high accuracy ($>93\%$). However, significant limitations persist: models struggle with false belief reasoning (19-83\% accuracy) and show high regional variance (20-30\% gaps). Crucially, ablation experiments 
reveal a modality gap, some frontier models exhibit a strong tendency to rely on parametric social priors, frequently defaulting to safety-aligned predictions even when visual context is removed.
We find that SOTA models show \textit{\textbf{social desirability bias}} - systematically favoring semantically positive answer choices over negative ones without visual evidence. Furthermore, while Chain-of-Thought prompting aids older models, it yields minimal gains for newer ones. Overall, our work provides a testbed for cross-cultural social reasoning, underscoring that despite architectural gains, achieving robust, visually grounded understanding remains an open challenge.
\end{abstract}

\section{Introduction}
Understanding human behavior in social contexts requires more than perceiving visible cues - it necessitates \textbf{Theory of Mind} (ToM), the cognitive ability to reason about others' beliefs, desires, intentions, and emotions, even when they remain hidden or must be inferred indirectly~\cite{premack1978does,baron1985does}.
In humans, ToM underpins essential social capacities including communication, coordination, empathy, and navigating cultural norms.
As artificial intelligence systems increasingly interact with people across diverse contexts, developing robust ToM reasoning capabilities has emerged as a critical frontier~\cite{kosinski2024theory,sap2022neural}.

Recent advances in vision-language models (VLMs) have yielded remarkable progress on image captioning, visual question answering, and grounded reasoning.
Current VLMs ~\cite{openai2023gpt4v, liu2024visual, team2023gemini, bai2023qwen, chen2024internvl} demonstrate strong performance when combining visual and linguistic understanding.
However, their capacity for \emph{social inference} - particularly in cultural contexts where mental states must be interpreted through the lens of cultural knowledge - remains far less explored. 

A gesture, ritual, or social role carries different mental-state implications across cultures.
\textit{For instance, sustained direct eye contact during conversation with an elder 
signals confidence and respectful engagement in many Western contexts, yet 
constitutes disrespect or inappropriate challenge in many East Asian, 
Indigenous, and Middle Eastern cultures, where averting one's gaze signals 
deference~\cite{mccarthy2006cultural}}. 
Without culturally grounded ToM reasoning, AI systems may misinterpret intentions, violate social expectations, or exhibit uneven behavior across cultural groups \cite{tao2024cultural}. Crucially, as models undergo safety alignment, a potential risk emerges: what we term an `illusion of empathy,' where systems may favor socially desirable or harmonious mental-state interpretations even when visual evidence is weak or absent \cite{sorin2024large}. Distinguishing true cross-cultural understanding from these alignment artifacts remains an under-addressed challenge.

\smallskip
\noindent\textbf{Background and Motivation.} While ToM has been studied extensively in cognitive science and increasingly in natural language processing, existing benchmarks reveal three critical limitations when considering VLMs operating across diverse cultural contexts.

\smallskip
\noindent\textbf{(a) Limited cultural scope in ToM benchmarks.}
In the language domain, ToMBench provides one of the most systematic evaluations of ToM capabilities in large language models, spanning 8 ToM tasks and 31 social-cognitive abilities~\cite{xie2024tombench}.
However, ToMBench focuses on textual scenarios without visual grounding and with limited cultural diversity.
In the multimodal space, MMToM-QA fuses textual and visual inputs for ToM tasks, yet its scenarios predominantly reflect Western everyday life - indoor settings, common objects, familiar social scripts - without systematic variation across cultural contexts~\cite{ma2024mmtom}.
More recently, \citet{wen2025howwell} propose an open-ended evaluation of VLMs on inferring human intentions from images using 30 carefully curated scenes
involving deception, conflicting interests, or hidden motives.
Nevertheless, their dataset size is modest and cultural diversity remains limited.
Embodied perspectives have also been explored: SoMi-ToM investigates agent-based social interaction in multimodal settings, examining first-person versus third-person perspective-taking~\cite{fan2025somi}.
More broadly, the existing ToM benchmarks rarely examine how \emph{culture itself} shapes the inference process - how rituals, symbols, dress codes, and social roles mediate our understanding of others' minds.

\smallskip
\noindent\textbf{(b) Underexplored cultural reasoning in VLMs.}
The capacity of VLMs to understand cultural context has recently attracted increased attention, though much of this work focuses on factoid-style comprehension rather than mental-state inference.
CulturalVQA, a benchmark drawn from 11 countries, probes VLM understanding of clothing, rituals, food, and traditions~\cite{nayak2024benchmarking}.
Their experiments reveal pronounced imbalances, with models performing substantially better on North American content than on other regions. CultureVLM extends this line of inquiry with a massive benchmark encompassing 19,682 cultural concepts across 188 regions, demonstrating that standard VLMs struggle significantly with non-Western content, and that targeted fine-tuning can partially - but not fully close the gap~\cite{liu2025culturevlm}.
CROPE examines how VLMs adapt to culture-specific concepts through in-context prompting, showing that effectively combining parametric knowledge with contextual cultural cues remains challenging~\cite{nikandrou2024crope}.
In a comprehensive survey of over 300 works, \citet{pawar2024survey} document how cultural awareness in multimodal and language models is often treated superficially or as an afterthought, with many systems failing to move beyond surface-level pattern matching.
This limitation extends to current benchmarks as well, which often test whether models can \emph{identify} cultural artifacts (naming a ritual, recognizing traditional dress) rather than whether they can perform \emph{mental-state inference grounded in cultural understanding}.

\smallskip
\noindent\textbf{(c) Methodological challenges.}
Building cross-cultural ToM benchmarks is hindered by expensive, slow manual annotation and biased, hallucinated model data that risks superficial, stereotypical, or derogatory reasoning  \cite{ignat2024annotations}. To be rigorous, these frameworks must navigate such hurdles to incorporate diverse sub-tasks across multiple cognitive complexity levels for nuanced social intelligence evaluation \cite{gandhi2023understanding}. 

\begin{figure*}[t]
    \centering
    \makebox[\linewidth][c]{%
        \begin{subfigure}[b]{0.32\linewidth}
            \centering
            \includegraphics[width=\linewidth]{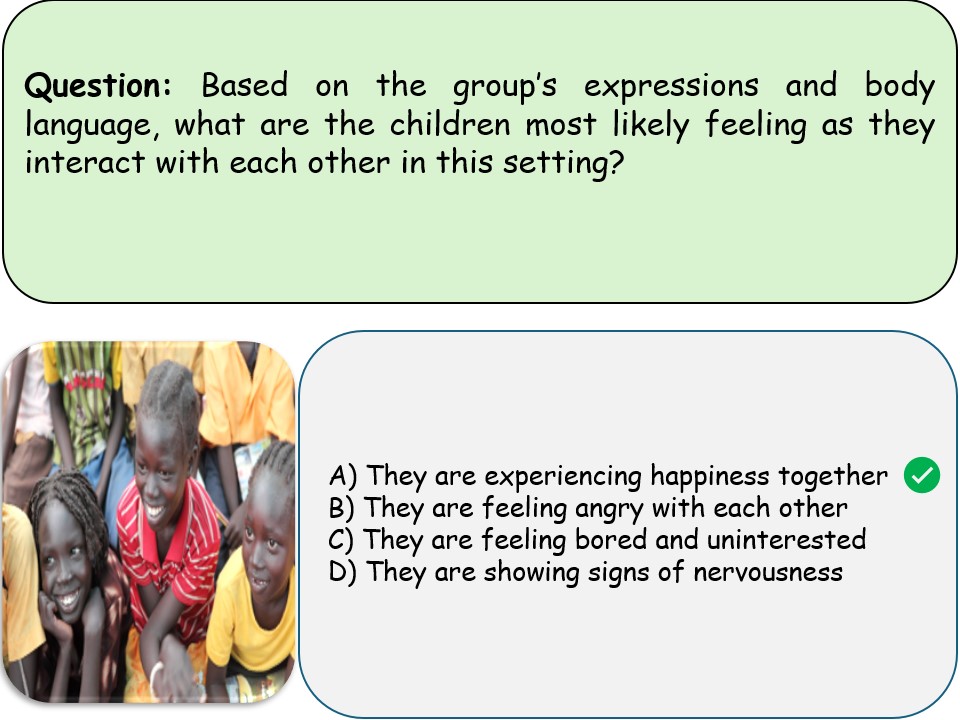}
            \label{fig:slide1}
        \end{subfigure}
        \hspace{1.5em}
        \begin{subfigure}[b]{0.32\linewidth}
            \centering
            \includegraphics[width=\linewidth]{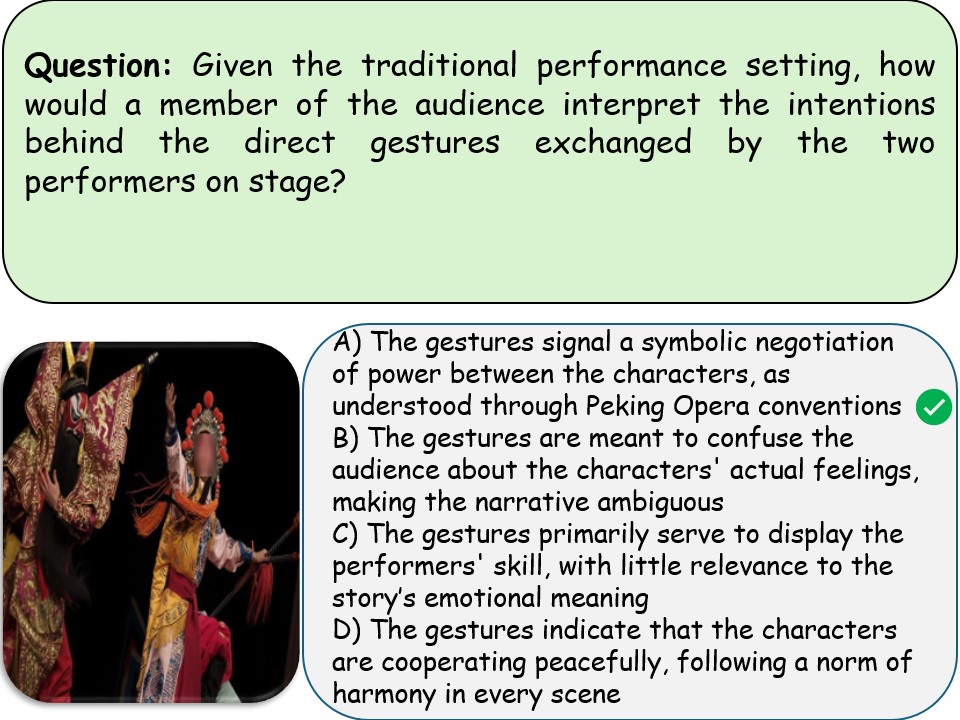}
            \label{fig:slide3}
        \end{subfigure}
    }
    \caption{Example questions from the CulturalToM-VQA dataset.}
    \label{fig:three_slides}
    \vspace{-1.5em}
\end{figure*}

\smallskip
\noindent\textbf{Our Contribution.}
We introduce \textbf{CulturalToM-VQA}, a benchmark comprising 5,095 question-answer pairs for cross-cultural, vision-based Theory of Mind tasks.
Our main contributions are:
\begin{itemize}[leftmargin=8pt, itemsep=0pt, topsep=0pt]
\item \textbf{Human-in-the-Loop (HITL) Annotation Pipeline:} We combine human expert curation of culturally rich imagery with GPT-assisted description and question generation, then validate outputs through human review following existing works \cite{kim2024meganno+, gandhi2023understanding}. This approach balances scalability with semantic reliability and cultural relevance.

\item \textbf{Structured ToM Task Taxonomy:} We formalize a taxonomy covering six ToM task types \textit{(mental state attribution, false belief reasoning, non-literal communication, social norm violations, perspective coordination, and multi-agent reasoning)} across four complexity levels, ranging from the perception of basic mental states to recursive reasoning about others’ beliefs, and extending to culturally grounded social coordination involving norms, intentions, and multi-agent interactions.

\item \textbf{Systematic Evaluation:} We systematically evaluate a broad set of VLMs using three prompting methods \textit{(zero-shot, zero-shot-CoT, compositional-CoT)}.  Importantly, \textit{our evaluation focuses on modern instruction-following multimodal LLMs rather than contrastive CLIP-style vision–language encoders, as the latter are not designed for open-ended or multiple-choice social reasoning.} We analyze both aggregate and stratified performance - across task types, complexity levels, data splits and examine fine-grained failure modes. 
\end{itemize}

We find a sharp capability gap across model generations: recent instruction-tuned VLMs achieve high accuracy on explicit ToM tasks (e.g, mental state attribution), yet all models degrade substantially on implicit reasoning such as \textit{false belief} and exhibit cross-cultural performance variance. We conduct supplementary no-image and option-level diagnostics on the \textit{best-performing models} to further interpret these results. The findings suggest a modest positivity heuristic in these models - likely reflecting safety alignment and socially normative priors - which partially explains no-image correctness, with remaining accuracy appearing to stem from higher-level semantic and social knowledge acquired during large-scale pretraining rather than simple lexical pattern matching \cite{strachan2024testing}.

\section{CulturalToM-VQA}
We construct CulturalToM-VQA via a three-stage pipeline: image selection by human annotators, proprietary MLLM GPT-assisted scene description with structured cues, and systematic question generation based on our task-complexity framework. Finally, we apply rigorous quality controls to ensure dataset integrity. We employ \textit{GPT-4.1} for annotation, crucially conditioning all prompts on the corresponding image to ensure generations are grounded in visual evidence rather than hallucinated priors. 
This methodology builds on evidence that GPT-4-class models serve as reliable human proxies for annotation \citep{jahan-etal-2024-finding, liyanage2024gpt} including ToM \cite{gandhi2023understanding} and effectively synthesize complex multimodal reasoning data \citep{liu2023visual, kim-etal-2023-fantom} . Some examples from the dataset are demonstrated in Figure \ref{fig:three_slides}.

\subsection{Data Sources}
We curate images for our dataset from the following three existing complementary image sources.

\smallskip
\noindent\textbf{(a) FindingEmo:} Unlike conventional emotion recognition datasets that focus on isolated faces or single individuals, FindingEmo \cite{mertens2024findingemo}  annotates 25{,}869 entire scenes collected from web depicting multiple people in naturalistic social settings making it particularly suitable for evaluating cultural ToM reasoning.

\smallskip
\noindent\textbf{(b) CulturalVQA:} \citet{nayak2024benchmarking} provides a culturally diverse benchmark of 2,378 image-question pairs sourced from 11 countries across 5 continents. While many questions focus on identifying cultural objects such as food, clothing, or artifacts, a selective subset involves humans or groups engaged in rituals and social activities that require contextual and inferential understanding. These instances, though fewer, are particularly relevant for evaluating ToM-related reasoning, as they demand recognition of culturally grounded intentions, shared beliefs, and social dynamics beyond surface-level visual cues.

\smallskip
\noindent\textbf{(c) CVQA:} \citet{romero2024cvqa} features a dataset of approximately 10,000 cross-cultural image–question pairs. The collection spans 30 countries across 4 continents and supports high linguistic diversity, covering 31 languages written in 13 distinct scripts. Although CVQA supports multiple languages, in our study, the evaluation is conducted in English.


\subsection{Image Selection}
Let $\mathcal{I}_{\text{source}}$ denote the union of images from our three source datasets. Human annotators performed a filtering operation to identify images suitable for Theory of Mind reasoning. Each image $I \in \mathcal{I}_{\text{source}}$ was evaluated according to four criteria: (i) social or mental state interpretability, defined as the presence of either multiple interacting agents or a single agent whose emotions, intentions, or social role can be inferred from contextual or cultural cues; (ii) observable interpersonal dynamics or role relationships; (iii) culturally specific contexts or practices; and (iv) emotional expressiveness conveyed through facial expressions or body language. Our final dataset comprises $\mathcal{I} = \{ I \in \mathcal{I}_{\text{source}} : I \text{ satisfies all four criteria} \}$, resulting in $|\mathcal{I}| = 394$ images after selection process (221 images from FindingEmo, 90 from CulturalVQA, and 83 from CVQA).


\subsection{Theory-of-Mind Task Taxonomy}
\paragraph{Task Description.} We select six abilities reflecting key stages in cognitive development: (i) \textit{Mental State Attribution (MSA)} refers to an individual's early-developing ability to interpret the observable actions and expressions of others (such as facial cues, body language, or vocal tone) to infer their underlying basic emotions and desires \citep{strachan2024testing,xie2024tombench}, (ii) \textit{False Belief Reasoning (FBR)} is the ability to attribute a mental state, specifically a false belief, to oneself or to another person, and to understand that this belief might differ from one's own correct understanding of reality\citep{kosinski2024evaluating}, (iii) \textit{Non-literal Communication (NLC)} is the ability to to understand messages where the intended meaning is not the same as the literal words spoken or written (such as interpreting irony, sarcasm, or indirect speech that requires higher-order intention tracking and pragmatic inference) \cite{winner1991distinguishing}, (iv) \textit{Social Norm Violations (SNV)} is the ability to perceive that a social error has occurred by integrating their knowledge of general social rules with an understanding of what the people involved in the situation actually know or believe \citet{baron1999recognition, strachan2024testing}, (v) \textit{Perspective Coordination (PC)} encompasses the progression from simple visual perspective-taking to reasoning about embedded and conflicting viewpoints \citet{frick2014picturing, wilf2024think}, and lastly, (vi) \textit{Multi-agent Reasoning (MAR)} involves tracking and reasoning on the most complex scenarios where multiple agents' beliefs and intentions must be tracked simultaneously \citet{kim2023fantom}. 

\begin{table}[h]
\centering
\small
\setlength{\tabcolsep}{6pt}
\resizebox{\columnwidth}{!}{
\begin{threeparttable}
\begin{tabular}{@{}C{5.2cm}C{0.9cm}C{0.9cm}C{0.9cm}C{0.9cm}@{}}
\toprule
\multicolumn{1}{@{}c@{}}{\raisebox{-0.6\normalbaselineskip}{\textsc{Task}}} &
\multicolumn{4}{c}{\textsc{Complexity Level}} \\
\cmidrule(l){2-5}
& \textbf{1} & \textbf{2} & \textbf{3} & \textbf{4} \\
\midrule
Mental State Attribution (MSA) & \textbf{1} & \textbf{1} & 0 & 0 \\
False Belief Reasoning (FBR)   & 0 & \textbf{1} & \textbf{1} & \textbf{1} \\
Non-literal Communication (NLC)& 0 & 0 & \textbf{1} & \textbf{1} \\
Social Norm Violations (SNV)   & 0 & 0 & \textbf{1} & \textbf{1} \\
Perspective Coordination (PC)  & 0 & \textbf{1} & \textbf{1} & \textbf{1} \\
Multi-agent Reasoning (MAR)    & 0 & 0 & 0 & \textbf{1} \\
\bottomrule
\end{tabular}
\end{threeparttable}}
\caption{\small Task–complexity Q-matrix, $\mathbf{Q} \in \{0,1\}^{6\times4}$ encoding the cognitive appropriateness of Theory-of-Mind tasks ($t_i$) at each complexity level ($c_j$). Bold entries denote $\mathbf{Q}_{ij}=1$, indicating that task $t_i$ is validly defined at complexity level $c_j$. 
}
\label{tab:qmatrix}
\vspace{-1.5em}
\end{table}

\paragraph{Task-Complexity Mapping.} We pair each ToM task with a graded complexity hierarchy which is required later to generate questions from images. We formalize this pairing with a Q-matrix from cognitive diagnosis models (CDMs), which encode item-skill relationships via an item-by-attribute incidence matrix \citep{qin2024using}. Let $\mathcal{T}=\{t_1,\dots,t_6\}$ be the set of ToM task types and let $\mathcal{C}=\{1,2,3,4\}$ denote four complexity levels. We define a binary matrix $\mathbf{Q}\in\{0,1\}^{|\mathcal{T}|\times|\mathcal{C}|}$ with entries
\[
\mathbf{Q}_{ij}=\begin{cases}
1 & \text{task } t_i \text{ is cognitively appropriate at } c_j,\\
0 & \text{otherwise.}
\end{cases}
\]
The set of admissible task--level pairs is $\mathcal{V}=\{(t_i,c_j)\in \mathcal{T}\times \mathcal{C}:\mathbf{Q}_{ij}=1\}$.
The Q-matrix ($\mathbf{Q}$)  as shown in Table \ref{tab:qmatrix} maps each Theory-of-Mind (ToM) task to the levels of cognitive complexity. These assignments are grounded in developmental psychology and pragmatics literature \citep{strachan2024testing, kosinski2024evaluating, winner1991distinguishing, baron1999recognition}. We provide detailed definitions and cognitive justifications for each specific task and mapping in Appendix~\ref{app:task_taxonomy}.


\subsection{Scene Description Generation}
For each image $I_i \in \mathcal{I}$ and metadata $M_i$, we use \textit{GPT-4.1} to generate structured annotations via a template $\mathcal{P}_{\text{scene}}$ (see Appendix~\ref{app:scene_prompt}) that prioritizes inferred mental states over surface-level description. The model produces a structured representation $\mathcal{D}_i = (d_i, E_i, T_i, C_i)$, where $d_i$ is a ToM-focused narrative, and $E_i, T_i, C_i$ are sets of phrases capturing emotional expressions, mental state inferences (beliefs, intentions), and cultural elements (symbols, rituals, norms) respectively.



\subsection{Structured Question Generation}

For each annotated image $I_i$ with structured description $\mathcal{D}_i=(d_i,E_i,T_i,C_i)$, we generate one multiple-choice question for each admissible task–complexity pair $(t,c)\in\mathcal{V}$ defined by the Q-matrix. Let $k$ index the set of valid task–complexity pairs for image $I_i$, i.e., $k=1,\dots,|\mathcal{V}|$. Each question is represented as:

{\small
\begingroup
\setlength{\abovedisplayskip}{2pt}
\setlength{\belowdisplayskip}{2pt}
\setlength{\abovedisplayshortskip}{2pt}
\setlength{\belowdisplayshortskip}{2pt}
\[
q_{i,k} = \big(s_{i,k},\, O_{i,k},\, a_{i,k},\, e_{i,k}\big),
\]
\endgroup
}

where $s_{i,k}$ is the question stem, $O_{i,k} = \{o_{i,1},o_{i,2},o_{i,3},o_{i,4}\}$ denotes the answer options, $a_{i,k} \in \{A,B,C,D\}$ identifies the correct answer, and $e_{i,k}$ provides a short explanation referencing supporting cues from $(E_i,T_i,C_i)$. For each valid $(t,c)$ pair, we ask \textit{GPT-4.1} with a structured prompt $\mathcal{P}_Q(t,c,\mathcal{D}_i)$ that inputs the narrative $d_i$ and cue sets to generate a question targeting the specific mental state construct $t$ at inference depth $c$. (See Appendix~\ref{app:mcq_prompt} for full templates).

\begin{figure*}[t]
    \centering
    
    \begin{subfigure}[b]{0.32\textwidth}
        \centering
        \includegraphics[width=\textwidth]{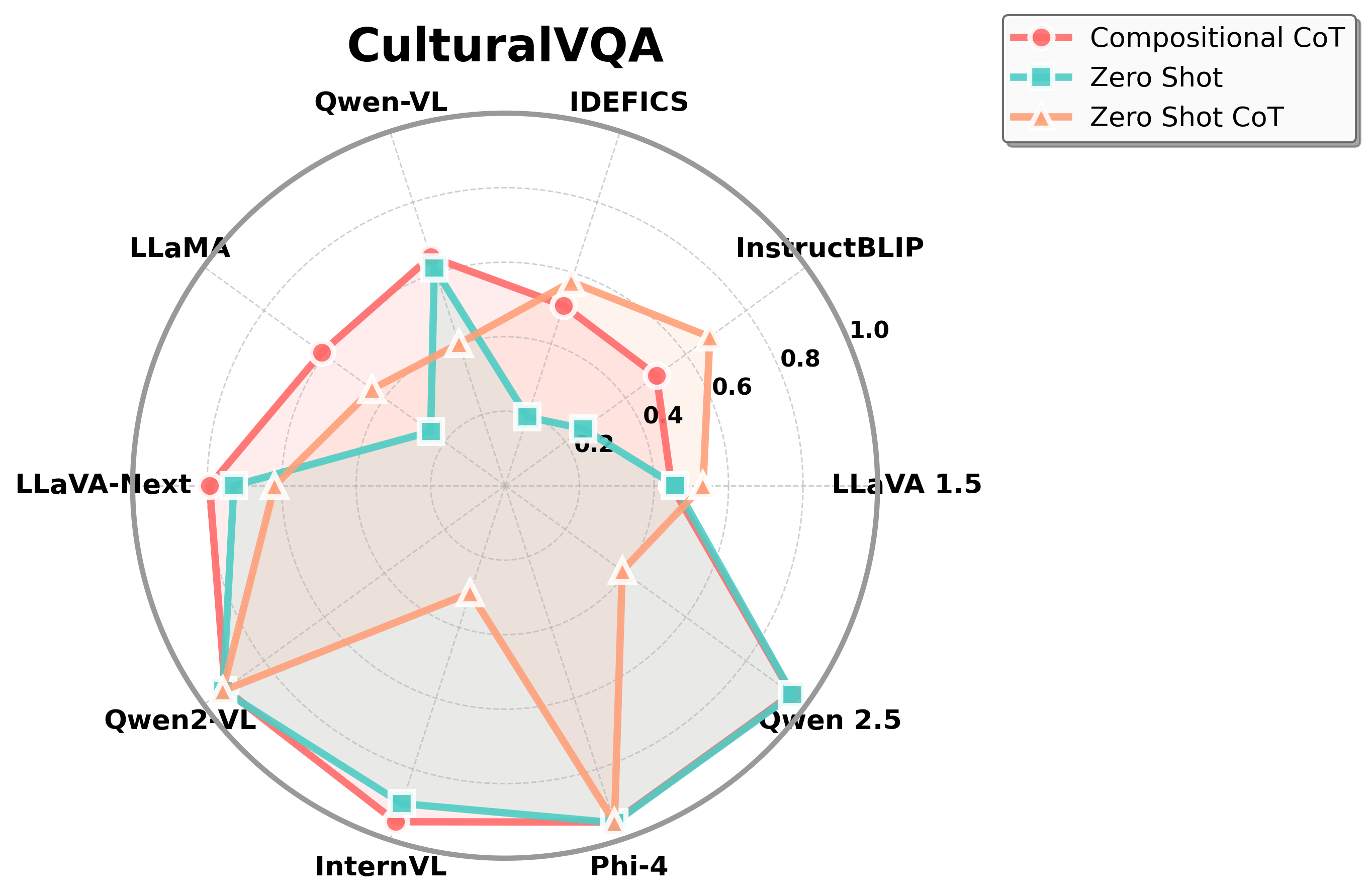}
        \caption{CulturalVQA Image Source}
        \label{fig:culturalvqa}
    \end{subfigure}
    \hfill
    \begin{subfigure}[b]{0.32\textwidth}
        \centering
        \includegraphics[width=\textwidth]{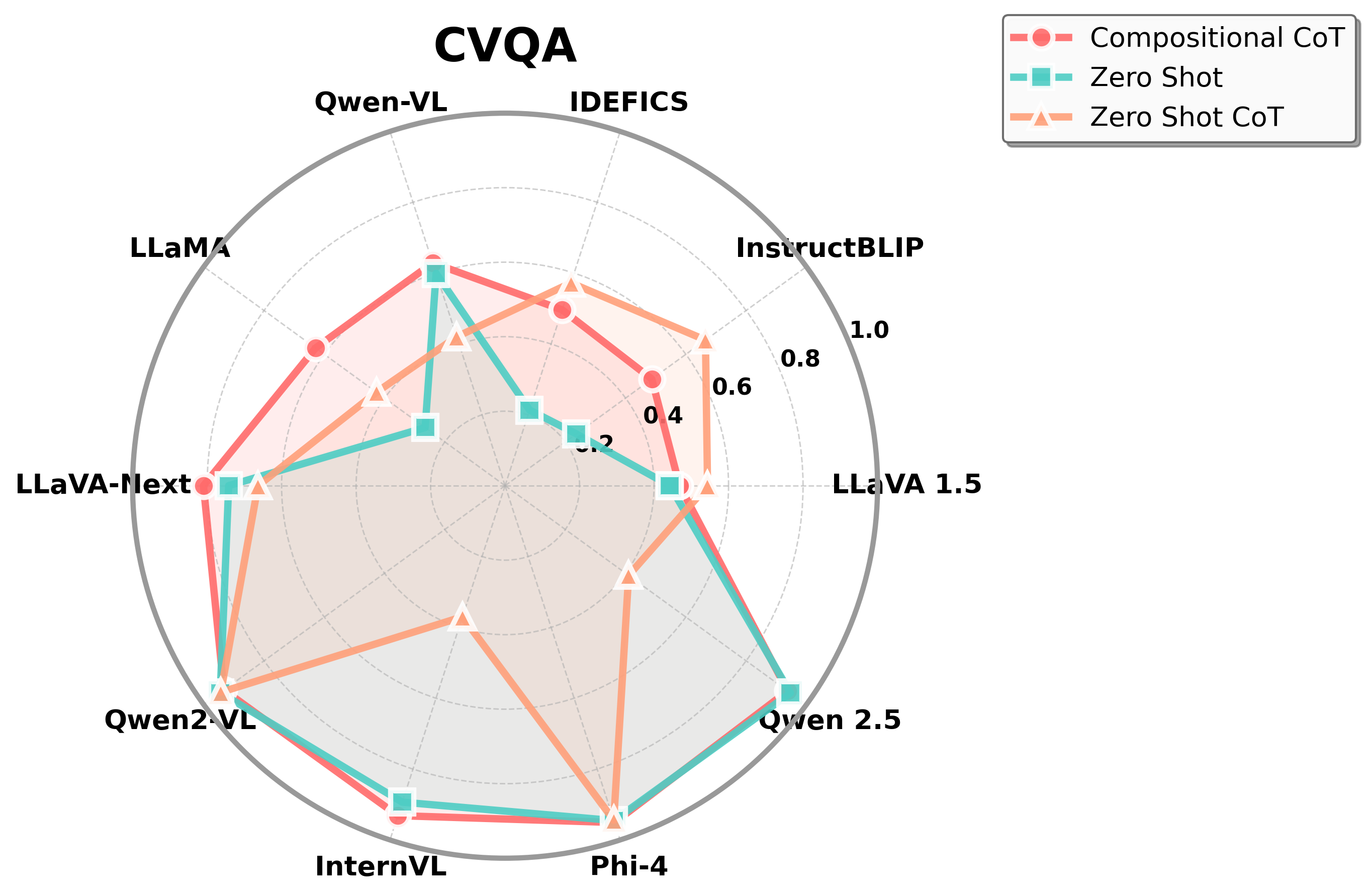}
        \caption{CVQA Image Source}
        \label{fig:cvqa}
    \end{subfigure}
    \hfill
    \begin{subfigure}[b]{0.32\textwidth}
        \centering
        \includegraphics[width=\textwidth]{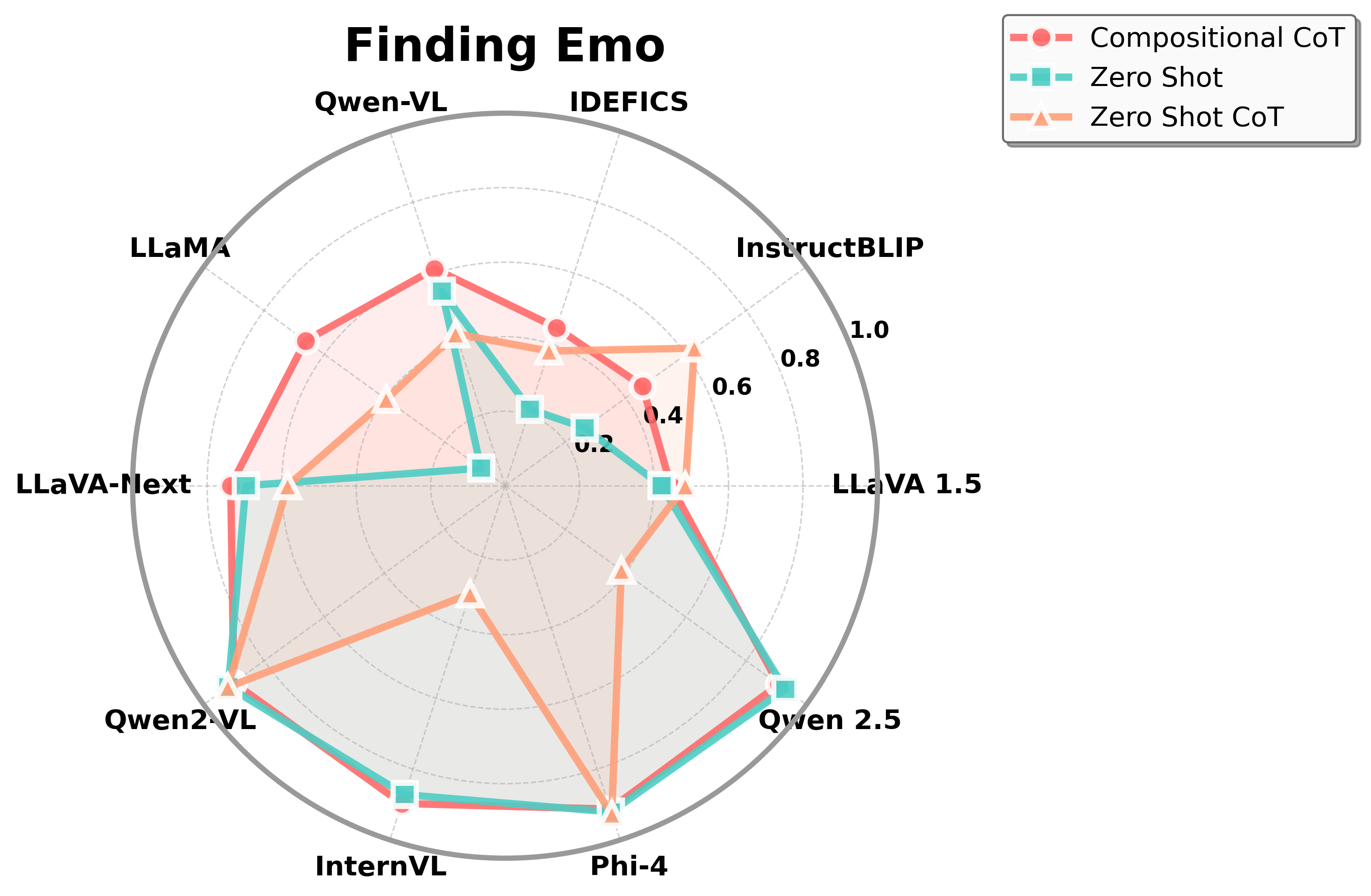}
        \caption{FindingEmo Image Source}
        \label{fig:finding_emo}
    \end{subfigure}
    
    \caption{Performance comparison of three prompting strategies across 10 vision-language models on images from three sources. Each radar chart displays 10 model axes with three overlaid polygons representing Compositional Chain-of-Thought (red circles), Zero Shot (teal squares), and Zero Shot CoT (orange triangles).}
    \label{fig:radar_comparison}
    \vspace{-1em}
\end{figure*}

\subsection{Validation and Randomization}
We apply a rigorous two-stage filtering: (1) an automated parser verifying structural integrity and uniqueness, and (2) expert review by two annotators with cultural studies backgrounds. Prior to validation, annotators completed a 40-hour training module on the specific cultural contexts and social norms represented in the images. Disagreements were adjudicated by 2 senior researchers. Finally, answer options were randomly permuted \textit{(seed 42)} to eliminate position bias. Additional details can be found in Appendix \ref{app:ans_randomization}.

\subsection{Dataset Statistics}
From an initial pool of $7,092$ generations across 394 images, we retained $N_{\text{valid}} = 5{,}095$ high-quality questions (on average $12.93$ questions per image) after validation. We report the task and level wise dataset distribution in Appendix \ref{app:dataset-composition}.

\subsection{Human Evaluation}
We assessed data quality on a stratified random sample ($N=533$, $\approx 10\%$). Two independent raters evaluated items across five dimensions, achieving substantial inter-rater reliability (Cohen's $\kappa$) \citep{cohen1960coefficient}. We observed high consensus on \textit{Cultural Appropriateness} ($\kappa=0.86$, 99.8\% agreement) and \textit{Complexity Correctness} ($\kappa=0.67$). Crucially, the \textit{Image Necessity} dimension ($\kappa=0.62$, 76.7\% agreement) confirmed that questions effectively test visual reasoning rather than textual priors. Additional details on human evaluation are provided in \ref{app:human_eval}.

\section{Experimental Details}

\smallskip
\noindent\textbf{(a) Evaluated Models.} We evaluated a diverse set of open-source vision–language models (VLMs) spanning different architectures, generations, and parameter scales. 
The selected models include InternVL-3.5-8B, LLaVA-1.5-7B, LLaVA-Next-7B, Qwen-VL Chat, Qwen2-VL-7B, Qwen2.5-VL-7B, Phi-4-6B, InstructBLIP-Vicuna, IDEFICS-9B, and Llama-11B; full model access links are provided in Appendix~\ref{app:model_urls}.  These models were chosen to capture a range of multimodal reasoning capabilities - from instruction-tuned LLM–VLM hybrids \textit{(e.g., LLaVA, InstructBLIP)} to more recent architectures optimized for high-resolution perception and advanced multimodal reasoning \textit{(e.g., Qwen2.5, InternVL-3.5)}.  All models were evaluated in their publicly released form using their respective HuggingFace checkpoints without further fine-tuning or hyperparameter modification.

\smallskip
\noindent\textbf{(b) Prompting Strategies.} Each VLM was prompted under three settings - \textit{zero-shot, zero-shot chain-of-thought,} and \textit{compositional chain-of-thought} - to assess differences in basic reasoning, step-by-step inference, and structured relational reasoning. All prompts follow a consistent structure across models for fair comparison. The complete templates are provided in Appendix~\ref{app:prompts}.

\smallskip
\noindent\textbf{Zero-shot prompting.}  
Following visual question answering evaluation from \citep{kojima2022large}, 
each model receives an image and a multiple-choice question containing four labeled options (\texttt{A–D}). The model is instructed to select the correct answer directly by outputting only the corresponding option letter. 

\smallskip
\noindent\textbf{Zero-shot CoT prompting.}  
We employ an extended variant, the zero-shot chain-of-thought prompting strategy introduced by \citet{kojima2022large}. The model is first prompted with ``\textit{Let's think step by step.}'' and required to conclude with an explicit answer format (``\textit{Answer: <option letter>}'').  

\smallskip
\noindent\textbf{Compositional CoT prompting.}  We adopt the two-stage compositional CoT framework proposed by \citet{mitra2024compositional}. In the first stage, the model is instructed to generate a JSON-formatted scene graph describing relevant \textit{objects, attributes,} and \textit{relations} observed in the image. In the second stage, the model uses this scene graph, along with the original question and image, to infer the answer. 

\smallskip
\noindent\textbf{(c) Evaluation Setup.}
We used default sampling parameters and native image preprocessing (RGB). The max token limit was set to 256, but increased to 512 for InternVL and Qwen2.5-VL in Zero-shot CoT to prevent truncation (see Appendix~\ref{app:zero_shot_cot_token_budget}). Zero-shot settings utilized a single inference pass, while Compositional CoT chained the intermediate scene graph to the reasoning prompt. Final predictions were extracted as the selected option.

\begin{table}[t]
\centering
\scriptsize
\setlength{\tabcolsep}{4pt}
\begin{threeparttable}
\begin{tabular}{@{}l l l c@{}}
\toprule
\textbf{Model} & \textbf{Rel. Date} & \textbf{Prompt Config.} & \textit{\textsf{Accuracy}} \\
\midrule
\multirow{3}{*}{InstructBLIP--Vicuna}
 & \multirow{3}{*}{May'23} & Zero-Shot         & \met{0.257} \\
 &                         & Zero-Shot CoT     & \met{0.648} \\
 &                         & Comp. CoT         & \met{0.473} \\
\midrule
\multirow{3}{*}{IDEFICS-9B-Instruct}
 & \multirow{3}{*}{Aug'23} & Zero-Shot         & \met{0.210} \\
 &                         & Zero-Shot CoT     & \met{0.466} \\
 &                         & Comp. CoT         & \met{0.471} \\
\midrule
\multirow{3}{*}{Qwen-VL-Chat}
 & \multirow{3}{*}{Aug'23} & Zero-Shot         & \met{0.575} \\
 &                         & Zero-Shot CoT     & \met{0.421} \\
 &                         & Comp. CoT         & \met{0.624} \\
\midrule
\multirow{3}{*}{LLaVA-1.5-7B}
 & \multirow{3}{*}{Sep'23} & Zero-Shot         & \met{0.433} \\
 &                         & Zero-Shot CoT     & \met{0.508} \\
 &                         & Comp. CoT         & \met{0.450} \\
\midrule[\heavyrulewidth]
\multirow{3}{*}{LLaVA-Next-7B}
 & \multirow{3}{*}{Jan'24} & Zero-Shot         & \met{0.713} \\
 &                         & Zero-Shot CoT     & \met{0.609} \\
 &                         & Comp. CoT         & \met{0.765} \\
\midrule
\multirow{3}{*}{Qwen2-VL-7B}
 & \multirow{3}{*}{Aug'24} & Zero-Shot         & \met{0.928} \\
 &                         & Zero-Shot CoT     & \met{0.929} \\
 &                         & Comp. CoT         & \met{0.914} \\
\midrule
\multirow{3}{*}{Llama-3.2-11B-Vision}
 & \multirow{3}{*}{Sep'24} & Zero-Shot         & \met{0.158} \\
 &                         & Zero-Shot CoT     & \met{0.412} \\
 &                         & Comp. CoT         & \met{0.643} \\
\midrule[\heavyrulewidth]
\multirow{3}{*}{Qwen2.5-VL-7B}
 & \multirow{3}{*}{Jan'25} & Zero-Shot         & \bestmet{0.939} \\
 &                         & Zero-Shot CoT     & \met{0.909}\textsuperscript{\hyperref[app:zero_shot_cot_token_budget]{*}} \\
 &                         & Comp. CoT         & \met{0.924} \\
\midrule
\multirow{3}{*}{Phi-4-multimodal}
 & \multirow{3}{*}{Mar'25} & Zero-Shot         & \met{0.933} \\
 &                         & Zero-Shot CoT     & \met{0.936} \\
 &                         & Comp. CoT         & \met{0.929} \\
\midrule
\multirow{3}{*}{InternVL-3.5-8B}
 & \multirow{3}{*}{Aug'25} & Zero-Shot         & \met{0.882} \\
 &                         & Zero-Shot CoT     & \met{0.891}\textsuperscript{\hyperref[app:zero_shot_cot_token_budget]{*}} \\
 &                         & Comp. CoT         & \met{0.916} \\
\bottomrule
\end{tabular}
\caption{Aggregate performance by model and prompting strategy.}
\label{tab:grouped_results}
\end{threeparttable}
\vspace{-1.7em}
\end{table}

\vspace{-18pt}

\begin{figure*}[ht]
    \centering
    \begin{subfigure}{0.49\textwidth}
        \centering
        \includegraphics[width=\textwidth]{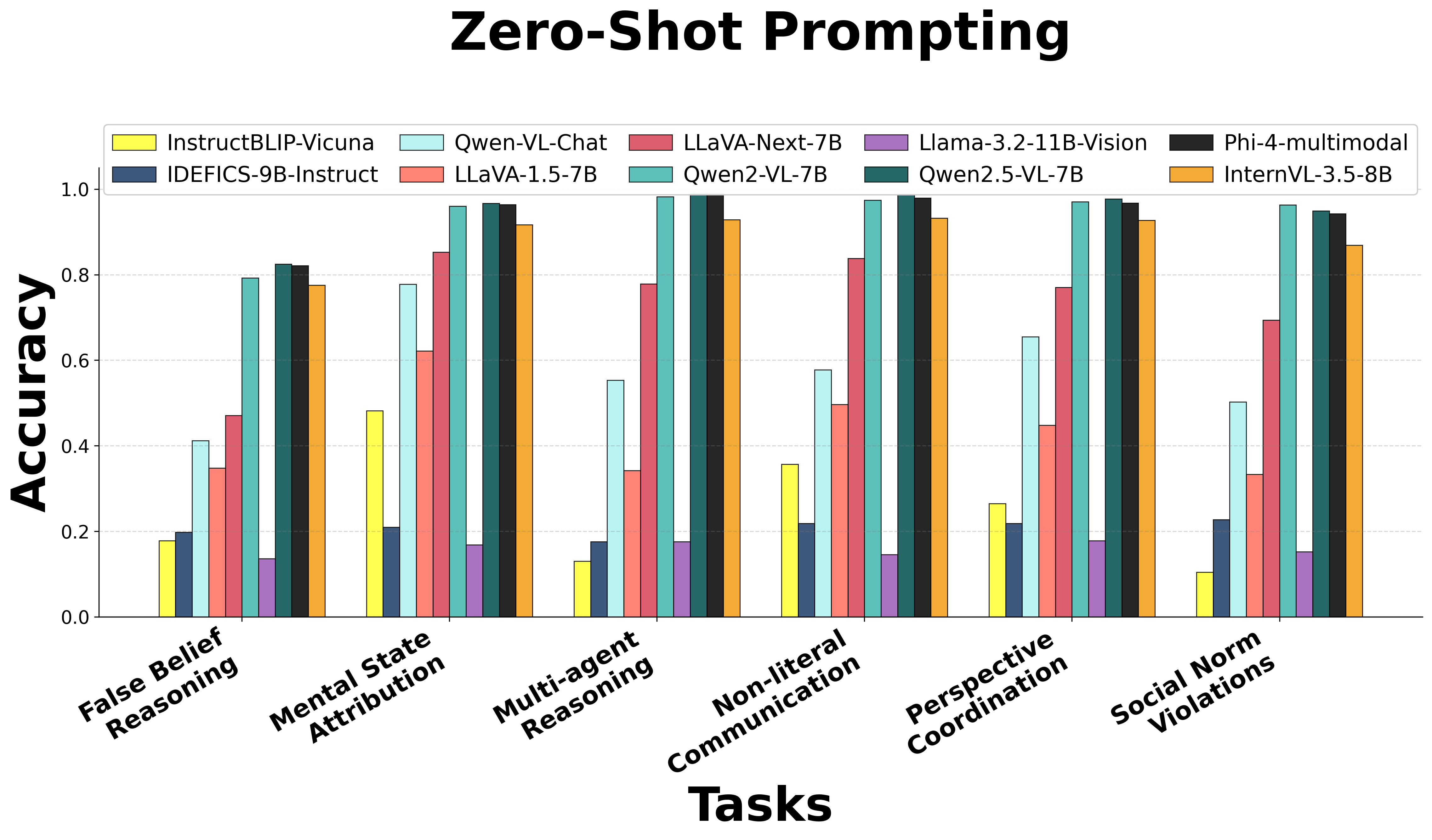}
        \label{fig:tom_zero_shot}
    \end{subfigure}
    \hfill
    \begin{subfigure}{0.49\textwidth}
        \centering
        \includegraphics[width=\textwidth]{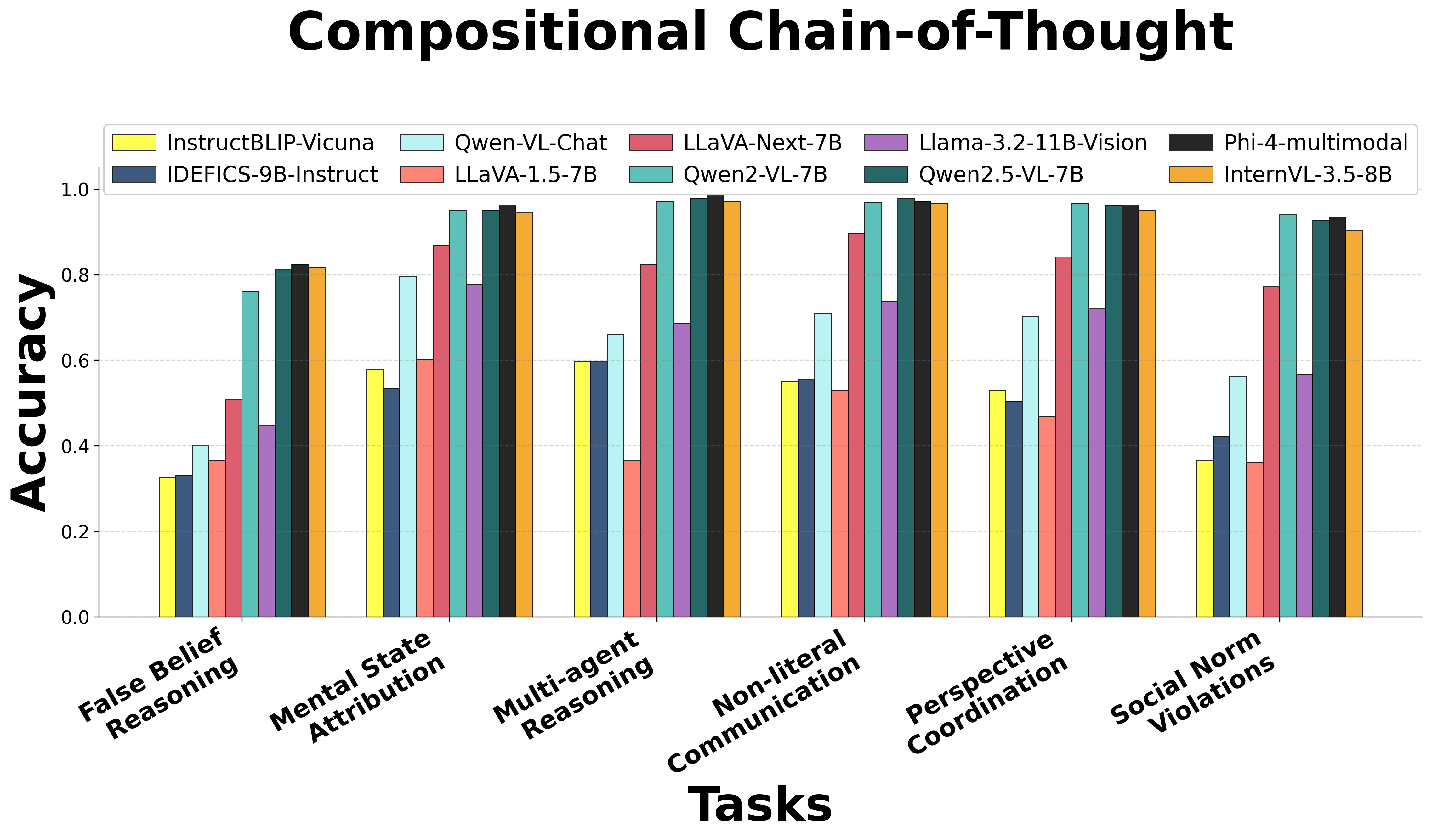}
        \label{fig:tom_comp_cot}
    \end{subfigure}
    
    \caption{Model performance across six Theory of Mind task categories with Zero Shot and Compositional CoT.}
    \label{fig:task_wise}
    \vspace{-1em}
\end{figure*}

\vspace{10pt}
\section{Results and Analysis}
\noindent\textbf{(a) Human Baseline.}
We evaluated human accuracy on a stratified sample of 481 questions balanced across task and complexity levels. Three independent raters achieved a majority-vote accuracy of \textbf{98.34\%} and a mean individual accuracy of 98.20\% ($\sigma = 0.0063$). This near-ceiling performance confirms the validity of the ground truth labels. Detailed sampling methodology and inter-rater analysis are provided in Appendix \ref{app:human_baseline}.

\smallskip
\noindent\textbf{(b) Overall Model Performance.} Table~\ref{tab:grouped_results} shows that Qwen-2.5-7B achieves the highest accuracy (\textbf{93.9\%}), followed closely by Phi-4-6B (93.6\%) and Qwen2-VL-7B (92.9\%). These recent architectures exhibit \textit{seemingly} strong performance with minimal sensitivity to prompting strategies. In contrast, earlier models like IDEFICS-9B (47.1\%) and InstructBLIP-Vicuna (64.9\%) perform significantly worse. Notably, while older and mid-tier models (e.g., LLaVA-Next, InternVL-3.5) benefit substantially from Compositional CoT (improving by 3–5\%), frontier models operate near a zero-shot ceiling, yielding negligible gains from explicit reasoning scaffolds.

\smallskip
\noindent\textbf{(c) Image source-wise Performance.} Figure~\ref{fig:radar_comparison} reveals a dramatic performance range, from 8.01\% (LLaMA-11B on FindingEmo images) to 95.38\% (Qwen 2.5 7B on CulturalVQA images). FindingEmo proves slightly more challenging (top scores 91--93\%) than CulturalVQA/CVQA (94--95\%), suggesting potential difficulty with web-sourced imagery. We observe a sharp generational gap: 2023 models (e.g., LLaVA 1.5) rarely exceed 68\%, while recent models (e.g., Qwen 2.5, Phi-4) consistently surpass 90\%, marking a 25--30\% improvement in one year. Regarding prompting, \textit{zero-shot CoT} yields inconsistent results; Table \ref{tab:vl_token_ablation_global_bold} in Appendix \ref{app:zero_shot_cot_token_budget} indicates that newer models require higher token budgets for effective reasoning. Conversely, \textit{compositional CoT} substantially benefits mid-tier and older models ($\approx 10\%$ gain), whereas frontier models like Phi-4 and Qwen2-VL maintain stability within 1.5\% variance across strategies.

\smallskip
\noindent\textbf{(d) Task-wise Performance.} Figure~\ref{fig:task_wise} reveals dramatic architectural progress: recent models (Qwen2-VL, Qwen2.5-VL, Phi-4, InternVL-3.5) achieve 82--99\% accuracy compared to legacy models (10--78\%). A notable exception is Llama-3.2-11B-Vision, which performs below 20\% despite its scale, mirroring early legacy models. Task difficulty varies substantially, with \textit{False Belief Reasoning} emerging as the most challenging task. We attribute this difficulty to the complex nature of identifying belief states in non-temporal data. In static VQA, False Belief Reasoning involves a disconnect between the observable reality shown in the image and the agent's internal knowledge. This requires identifying cues indicating that the agent perceives the scene differently than the omniscient viewer. Unlike video-based tasks showing belief formation, we treat static False Belief as a multimodal integration challenge: the image provides the visual reality, while the question establishes a limitation in the agent's knowledge. To answer correctly, the model must suppress its reliance on the visible ground truth and deduce the agent’s mistaken belief based on the textual constraints - an abstract reasoning step significantly harder than direct visual recognition. While False Belief and Social Norm Violations remain difficult, other tasks prove more tractable.

\begin{figure}[h]
    \centering
    \includegraphics[width=\columnwidth]{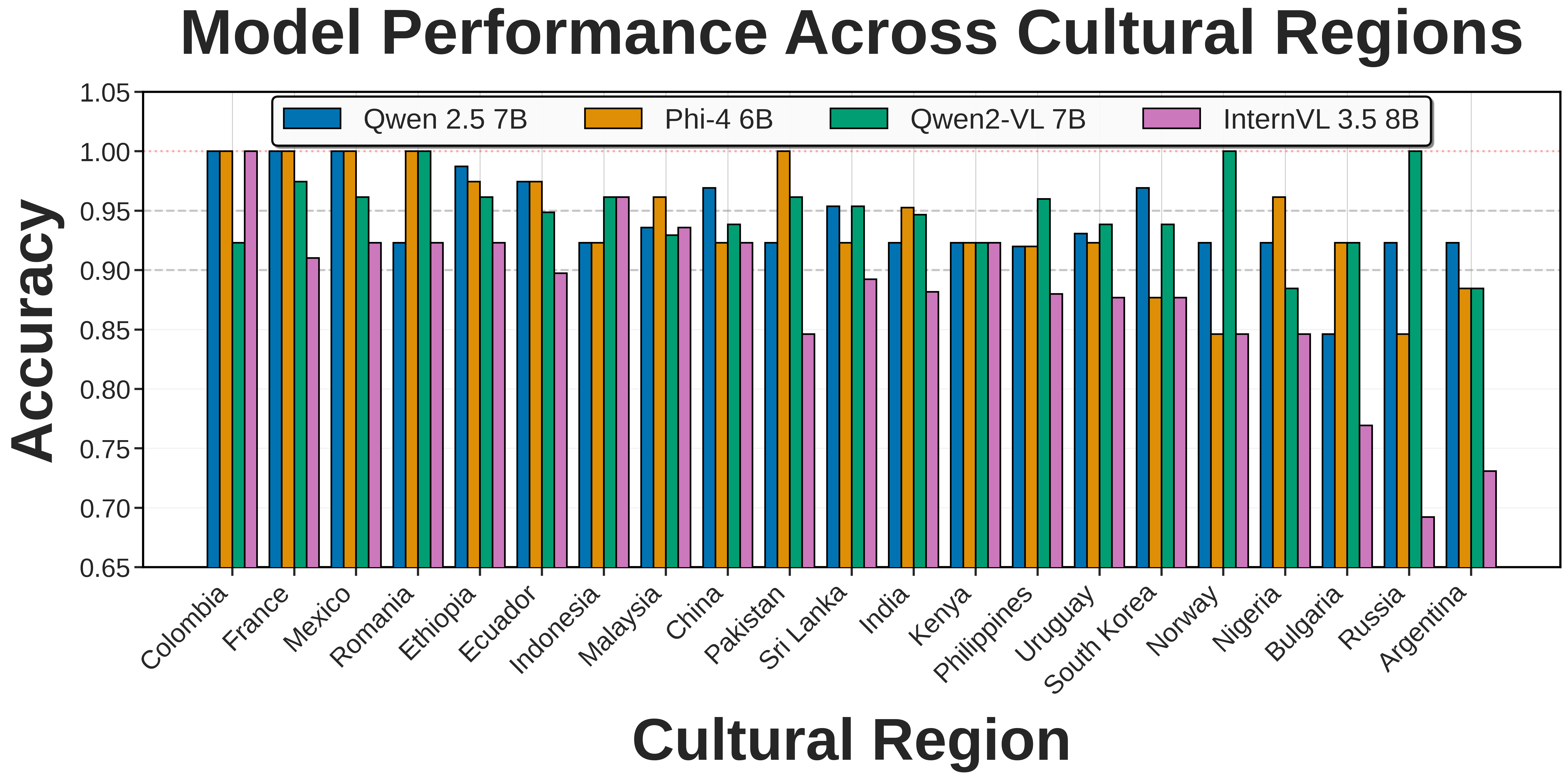}
    \caption{Model performance across cultural regions on CVQA images show high variance. Existing meta-data from CVQA dataset was used for regions.}
    \label{fig:cross-cultural-acc-cvqa}
    \vspace{-1.5em}
\end{figure}

\noindent\textbf{(e) Complexity-wise Performance.} Performance across four complexity levels in Figure~\ref{fig:complexity_wise} of Appendix shows a clear capability hierarchy, with modern models (right side of plots) achieving near-perfect accuracy on Complexity 1 (97--100\%) while legacy models (left side) struggle even at basic reasoning. All models exhibit graceful degradation through Complexity 2--3, but Complexity 4 represents a universal challenge where even top performers drop to 56--87\% in zero-shot settings - a major decline from Complexity 3. Compositional CoT prompting proves particularly valuable at Complexity 4, providing improvement across all model tiers, with the largest gains for older VLMs. This suggests that while modern architectures show good performance on lower complexity tasks, highly complex cross-cultural multi-step scenarios still remain a fundamental challenge.

\smallskip
\noindent\textbf{(f) Cross-Cultural Bias Analysis.} Figure \ref{fig:cross-cultural-acc-cvqa} reveals systematic zero-shot performance disparities across cultural regions on the CVQA split our dataset, proving that VLMs lack cultural uniformity. While regions like Colombia, France, Mexico, and Romania show consistent high accuracy ($>95\%$), Argentina and Russia see significantly lower performance, particularly for InternVL-3.5-8B, which drops to 70--75\%. InternVL underperforms generally, with drastic drops in Bulgaria, Russia, and Argentina. In contrast, Qwen2-VL-7B achieves perfect accuracy in Norway and Russia where others struggle. Phi-4-6B and Qwen-2.5-7B remain largely stable but dip in Russia and Bulgaria. Tight clustering in Colombia and France suggests these contexts are well-represented, whereas Norway, Nigeria, Bulgaria, and Russia show large gaps between models, highlighting that ``cross-cultural'' capability varies heavily by architecture and training distribution.

\section{Disentangling Visual Grounding from Alignment Priors}
\label{sec:alignment_analysis}
To rigorously assess whether high VLM performance stems from genuine visual perception anc culural understanding or language priors, we conducted a visual ablation study. We focus our analysis on the Qwen family as it uniquely spans the trajectory from early multimodal capabilities (Qwen-VL: 57\% accuracy) to best overall performance (Qwen2.5: 94\% accuracy), allowing us to trace the emergence of non-visual reasoning patterns.

\begin{table}[h]
\centering
\tiny
\resizebox{\columnwidth}{!}{
\begin{tabular}{l l cc cc}
\toprule
\multirow{2}{*}{Image Source} & \multirow{2}{*}{Model} 
& \multicolumn{2}{c}{Parametric Prior (Blind)} 
& \multicolumn{2}{c}{Visually Grounded Inference} \\
& & Acc. & Macro F1 & Acc. & Macro F1 \\
\midrule

\multirow{3}{*}{\textbf{CulturalVQA}} 
& Qwen-VL-Chat        & 0.3978 & 0.3730 & 0.6092 & 0.4879 \\
& Qwen2-VL-7B         & 0.9350 & 0.9351 & 0.9359 & 0.9358 \\
& Qwen2.5-VL-7B       & 0.9462 & 0.9460 & \textbf{0.9573} & \textbf{0.9572} \\
\midrule

\multirow{3}{*}{\textbf{CVQA}} 
& Qwen-VL-Chat        & 0.3910 & 0.2849 & 0.6171 & 0.4065 \\
& Qwen2-VL-7B         & 0.9416 & 0.9414 & \textbf{0.9453} & \textbf{0.9452} \\
& Qwen2.5-VL-7B       & 0.9397 & 0.9395 & 0.9416 & 0.9339 \\
\midrule

\multirow{3}{*}{\textbf{FindingEmo}} 
& Qwen-VL-Chat        & 0.3731 & 0.2295 & 0.5591 & 0.3702 \\
& Qwen2-VL-7B         & 0.9048 & 0.9048 & 0.9185 & 0.9185 \\
& Qwen2.5-VL-7B       & 0.9048 & 0.9049 & \textbf{0.9294} & \textbf{0.9291} \\
\bottomrule
\end{tabular}}
\caption{Zero-shot Accuracy and Macro-F1 of Qwen vision-language models with and without visual input on cross-cultural Theory-of-Mind.}
\label{tab:qwen_vl_with_without_image}
\vspace{-1.5em}
\end{table}

\smallskip
\noindent\textbf{(a) The `Blind' Baseline.} From Table~\ref{tab:qwen_vl_with_without_image}, we find that while the older Qwen-VL-Chat drops to near-random performance without visual context ($\approx 39\%$), Qwen2-VL and Qwen2.5-VL maintain high accuracy. All models are benefited with visual context overall. This suggests that some frontier models can effectively shortcut visual reasoning, leveraging parametric social priors learnt during training to approximate the correct answer. We investigate two hypotheses for this behavior: (1) simple lexical leakage, or (2) safety-aligned social priors.

\smallskip
\noindent\textbf{(b) Excluding Lexical Shortcuts.} We first test if questions contain exploitable keyword patterns. We trained a TF-IDF + Logistic Regression baseline under cross-validation. Across all splits, this baseline performs near chance ($\approx 25\%$; e.g., $0.233 \pm 0.02$ on \textit{CulturalVQA}), confirming that the high accuracy of frontier models is \textbf{not} driven by surface-level lexical heuristics. Additional details can be found in Appendix \ref{app:tfidf_lr}.

\smallskip
\noindent\textbf{(c) Quantifying Social Desirability Bias.} We next investigate if safety alignment drives models to heuristically select "positive" or "safe" answers. We utilize a sentiment-based "Positivity Heuristic" classifier that selects the most semantically positive option.

\smallskip
\noindent\textbf{Dataset Bias:} A simple positivity heuristic yields accuracies between $45\%$ (\textit{FindingEmo}) and $56\%$ (\textit{CulturalVQA}). This reflects our rigorous exclusion of culturally offensive or stereotypical content during annotation, which naturally skews the ground truth toward socially constructive interactions.

\smallskip
\noindent\textbf{Model Bias:} 
When VLMs make incorrect predictions, they are more likely than chance to select an option that is more positive than the correct answer - a behavior we quantify using \textit{Excess Positivity Drift}. 
For Qwen2.5-VL on \textit{CulturalVQA}, this drift is substantial ($0.111$, $p<0.05$), corresponding to an 11 percentage-point increase over chance among distractors.
This tendency is further amplified in \textit{conflict} cases, where the correct answer is not the most positive option: in such cases, the model selects the most positive distractor 22 percentage points more often than expected under random choice.
Appendix \ref{app:qual_text_only} provides additional qualitative evidence showing that, without visual input, models often default to socially favorable interpretations (e.g., happiness or comfort) in ambiguous scenarios instead of abstention.

While social desirability bias explains a substantial portion of the performance ($\approx 56\%$), it does not account for the full $>93\%$ accuracy. Since simple lexical patterns are ruled out, we conclude that the remaining performance gap stems from the models' extensive pre-trained world knowledge, enabling them to infer plausible social contexts from text alone, independent of visual grounding.

\section{Conclusion}

We introduced \textbf{CulturalToM-VQA} to evaluate visually grounded Theory of Mind across diverse cultures. Benchmarking VLMs reveals a sharp generational shift, with frontier models achieving near-human accuracy on explicit tasks. However, critical gaps persist: models struggle with false belief reasoning and exhibit inconsistent generalization across cultural regions. Crucially, visual ablation exposes a visual insensitivity where SOTA models maintain high performance without images. We trace this to pre-trained world knowledge and social desirability bias, where safety alignment predisposes models toward ``positive'' answers when grounding is absent. These findings indicate that while architecture and alignment have enhanced surface-level reasoning, achieving genuine, culturally informed Theory of Mind remains an open challenge for multimodal AI.

\section{Limitations}

\paragraph{Visual Necessity and Parametric Knowledge.}
A primary limitation of this work is that strong performance on CulturalToM-VQA does not uniformly imply strict visual grounding. Visual ablation reveals a visual bypass tendency, where some frontier VLMs effectively shortcut explicit visual processing by leveraging parametric priors to approximate socially aligned answers. At the same time, we observe a clear generational distinction: older VLMs exhibit substantially lower overall accuracy and show markedly larger performance gains when visual input is available, indicating a stronger dependence on visual grounding. Accordingly, CulturalToM-VQA should be interpreted as a diagnostic instrument for examining how different generations of models integrate visual and social information, rather than as a definitive test that correct answers always require visual perception.

\paragraph{Ambiguity in Image Necessity for Social Reasoning.}
Human evaluation reveals only moderate agreement on whether an image is strictly required to answer a given question. This reflects a broader challenge inherent to Theory of Mind tasks, where mental-state inference often combines visual cues with commonsense and culturally shared knowledge. Consequently, many questions are not designed to make the image the sole source of information, but rather to situate the social interaction within a culturally meaningful visual context. As such, image necessity should be understood as contextual rather than absolute.

\paragraph{Dataset Scale and Inter-Question Dependence.}
Although CulturalToM-VQA contains over 5,000 questions, these are derived from a limited set of curated images, with multiple questions associated with each image. This design introduces dependencies that may allow models to exploit recurring scene archetypes or culturally familiar scripts, potentially inflating performance without demonstrating robust generalization. While this structure enables controlled variation across Theory of Mind task types and complexity levels, it constrains claims about performance on entirely novel visual contexts.

\paragraph{Cultural Coverage and Language Constraints.}
While the benchmark targets culturally diverse social scenarios, all questions and evaluations are conducted in English. As a result, the dataset evaluates culturally grounded content rather than cross-lingual or language-specific Theory of Mind reasoning. Performance should therefore not be interpreted as evidence of multilingual cultural competence.

\paragraph{Dataset Construction and Model Involvement.}
CulturalToM-VQA is constructed using a frontier proprietary multimodal language model, followed by extensive expert human verification. Although this process substantially improves quality and consistency, it may introduce stylistic regularities or biases characteristic of contemporary large language models. Consequently, the benchmark is not fully independent of current model-generation paradigms, and future extensions would benefit from alternative collection or generation strategies.

\paragraph{Potential Training Data Exposure.}
Because the source images originate from publicly available datasets, it is possible that some evaluated models were exposed to related visual content during pretraining. While our visual ablations and baseline comparisons help contextualize performance, we cannot fully rule out indirect memorization at the image or scene level.

\paragraph{Non-Orthogonality of Theory of Mind Tasks.}
Although questions are categorized into distinct Theory of Mind task types and complexity levels, these dimensions are not strictly orthogonal. For example, false-belief reasoning may implicitly require perspective coordination or emotion recognition. The proposed taxonomy should therefore be viewed as a practical analytical framework rather than a strict cognitive decomposition.

\paragraph{Intended Use of the Benchmark.}
Taken together, these limitations indicate that CulturalToM-VQA is most appropriately used as a structured analytical benchmark rather than as a single-score evaluation. Its primary purpose is to facilitate fine-grained analysis of when and how vision--language models succeed, fail, or rely on unintended shortcuts in culturally grounded social reasoning, rather than to provide a definitive measure of general Theory of Mind capability.
\section{Ethics Statement}
\textbf{Cultural Essentialism and Stereotyping.} A primary ethical concern in evaluating cross-cultural Theory of Mind (ToM) is the risk of reducing complex, fluid cultural dynamics to static or essentialized representations. By employing a VLM-assisted pipeline to extract ``cultural cues'' and generate questions anchored in specific traditions, there is an inherent risk that the model may reinforce existing stereotypes or produce reductive generalizations about non-Western cultures. We acknowledge that culture is not monolithic; rituals, gestures, and social norms vary significantly by region, sub-group, and context. To mitigate this, we utilized a human-in-the-loop verification process where annotators with backgrounds in cultural studies reviewed items for cultural appropriateness, achieving high agreement ($\kappa=0.86$) on the cultural fidelity of the content. However, users of this dataset should remain aware that the ``correct'' reasoning reflects a consensus between expert verification and model generation, rather than an absolute ground truth of the lived cultural experience.

\textbf{Bias in Model Generated Annotations.} We explicitly state that relying on GPT-4.1 for data generation introduces the risk of propagating the model's intrinsic training biases. Large language models have been shown to exhibit Western-centric biases in social reasoning and may hallucinate cultural details when parametric knowledge is sparse. While our pipeline incorporated strict "grounding" constraints - requiring questions to rely on observable visual cues with expert supervision rather than external priors  - the resulting dataset inevitably reflects the cultural knowledge boundaries of the teacher model. We caution against using high performance on this benchmark as definitive proof of a model's cultural competence in real-world deployments, as models may still struggle with embodied cultural interactions not captured by static QA pairs.

\textbf{Privacy and Data Usage.} The images included in CulturalToM-VQA are curated from three existing public datasets: FindingEmo, CulturalVQA, and CVQA. We adhered to the usage policies of these source datasets. In our selection of the images, we manually filtered scenes to ensure they do not depict individuals in compromising, undignified, or harmful situations and we do not claim ownership of the source images. 


\textbf{Broader Impact and Dual Use.} Developing AI systems capable of inferring hidden mental states and cultural intentions carries significant dual-use risks. While robust cross-cultural ToM is essential for empathetic human-AI interaction and reducing bias in automated systems, these capabilities could theoretically be misused for manipulative social engineering or surveillance that exploits cultural vulnerabilities. By highlighting the current limitations of VLMs - specifically their struggle with False Belief Reasoning and Social Norm Violations - we aim to steer research toward more robust and transparent social reasoning, rather than deceptive applications.

\bibliography{custom}

\appendix

\section*{Appendix}
\label{sec:appendix}

\section{Dataset Distribution}
\begin{figure*}[t]
    \centering
    \begin{subfigure}[b]{0.52\linewidth}
        \centering
        \includegraphics[width=\linewidth]{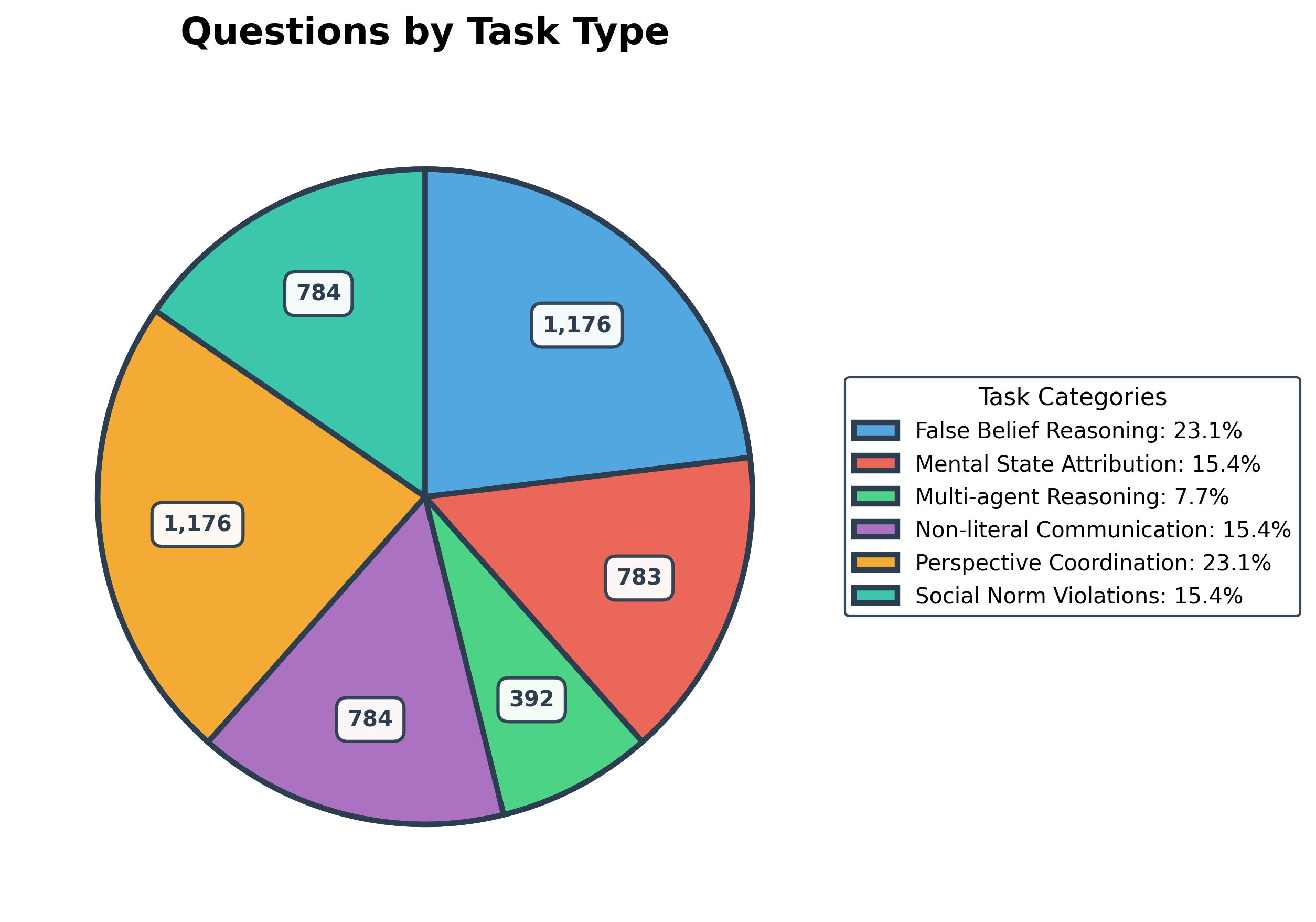}
        \caption{Distribution across six task categories}
        \label{fig:task}
    \end{subfigure}
    \hfill
    \begin{subfigure}[b]{0.45\linewidth}
        \centering
        \includegraphics[width=\linewidth]{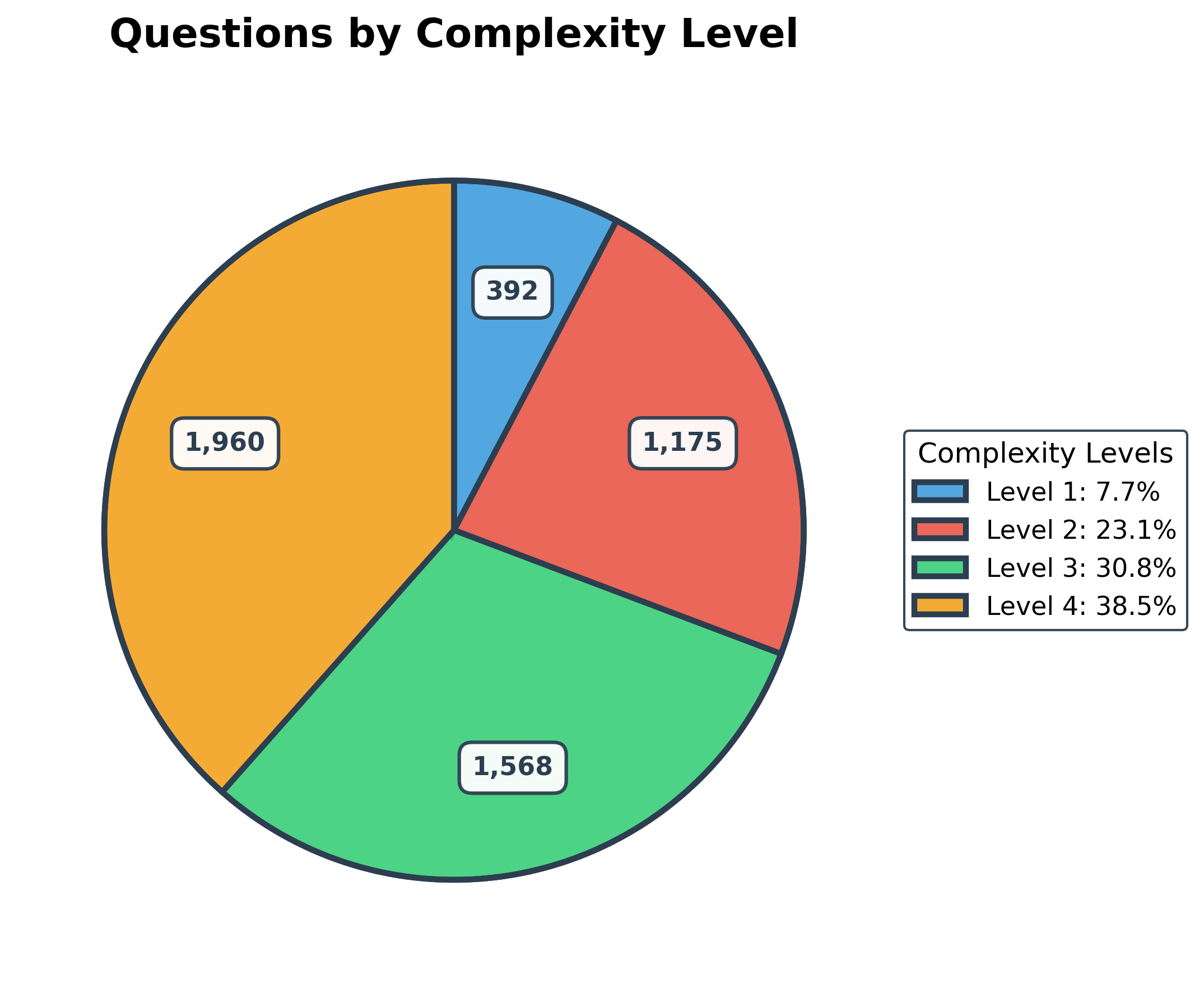}
        \caption{Distribution across four complexity levels}
        \label{fig:complexity}
    \end{subfigure}

    \caption{Dataset distribution with $N=5{,}095$ questions. (a) Coverage across six cognitive task types with counts shown in each slice. (b) Progression from complexity level~1 to~4.}
    \label{fig:dataset}
\end{figure*}

The CulturalToM-VQA dataset comprises a total of 5,095 questions spanning diverse cognitive dimensions. As shown in Figure~\ref{fig:dataset}, the task distribution is led by False Belief Reasoning and Perspective Coordination, followed by Mental State Attribution, Non-literal Communication, Social Norm Violations, and Multi-agent Reasoning. Similarly, the complexity distribution is skewed towards higher-order reasoning, with Level 4 comprising the largest share, followed in descending order by Level 3, Level 2, and Level 1.
\label{app:dataset-composition}

\section{Theory-of-Mind Task Taxonomy}
\label{app:task_taxonomy}

We include six Theory of Mind (ToM) abilities chosen because they reflect key stages in human cognitive development, capture both social and pragmatic aspects of reasoning, and show clear differences in difficulty for large language models.
The Q-matrix ($\mathbf{Q}$)  as shown in Table \ref{tab:qmatrix} maps each Theory-of-Mind (ToM) task to the levels of cognitive complexity at which it is meaningfully expressed, guided by developmental evidence and empirical results from LLM evaluations. 
\textit{Mental State Attribution (MSA)} spans Levels~1–2, reflecting early-emerging abilities to infer basic emotions and desires from observable cues \citep{strachan2024testing,xie2024tombench}. 
\textit{False Belief Reasoning (FBR)} is associated with Levels~2–3, capturing the developmental shift from understanding others’ beliefs to reasoning about second-order beliefs \citep{kosinski2024evaluating}. 
\textit{Non-literal Communication (NLC)} corresponds to Levels~3–4 \cite{winner1991distinguishing}, as interpreting irony, sarcasm, or indirect speech requires higher-order intention tracking and pragmatic inference. 
\textit{Social Norm Violations (SNV)} also align with Levels~3–4, since detecting faux pas or inappropriate actions requires integrating knowledge of social conventions with mental-state attribution \citet{baron1999recognition, strachan2024testing}. 
\textit{Perspective Coordination (PC)} extends across Levels~2–4, encompassing the progression from simple visual perspective-taking to reasoning about embedded and conflicting viewpoints \citet{frick2014picturing, wilf2024think}. 
Finally, \textit{Multi-agent Reasoning (MAR)} is limited to Level~4, representing the most complex scenarios where multiple agents’ beliefs and intentions must be tracked simultaneously \citet{kim2023fantom}.

\begin{enumerate}[leftmargin=1.35em,itemsep=2pt,topsep=2pt]
\item \textbf{Task $t_1$: Mental State Attribution (MSA).} This task involves recognizing emotions, desires, and knowledge states from observable behavior such as facial expressions, gaze, or posture. 
It represents the foundational Theory of Mind (ToM) ability on which higher-order reasoning builds. 
Both humans and current large language models (LLMs) show near-ceiling performance on simple attribution tasks \cite{strachan2024testing, xie2024tombench}. In this benchmark, Mental State Attribution strictly necessitates visual processing to identify affective cues (e.g., facial expressions or posture), which serve as the raw evidence that must be mapped to mental states within the specific cultural context provided.

\item \textbf{Task $t_2$: False Belief Reasoning (FBR).} 
This task involves understanding that an agent can hold a belief that differs from reality, allowing one to distinguish between mental representations and the actual state of the world. 
False belief reasoning is regarded as a core stage in the development of Theory of Mind, typically emerging around ages 4–5 when children begin to predict actions based on others’ mistaken beliefs rather than factual conditions. GPT-4 achieves approximately 75\% accuracy on standard false belief tasks, comparable to the performance of 6-year-old children but still below adult reasoning levels \cite{kosinski2024evaluating}. In our benchmark, these tasks necessitate multimodal integration: the textual scenario establishes the specific cultural or situational context - often containing the agent's true intent or limitations in knowledge - while the visual input provides the observable behavioral cues that differ from the agent's internal belief, forcing the model to distinguish between the visual appearance and the true mental state.

\item \textbf{Task $t_3$: Non-literal Communication (NLC).} This task encompasses understanding irony, sarcasm, hints, and indirect requests, which typically requires attributing second-order intentions - recognizing that a speaker intends the listener to recognize that the literal utterance is not the intended meaning. \cite{winner1991distinguishing} demonstrated that irony comprehension requires recognising a speaker’s second‐order intention, that is, understanding what the speaker intends the listener to believe about their intended meaning. In Non-literal Communication tasks in the benchmark, the image generally provides the physical instantiation of a communicative act which must be visually recognized before it can be mapped to its symbolic second-order meaning using the specific cultural norms defined in the question stem. 

\item \textbf{Task $t_4$: Social Norm Violations (SNV).}
This task involves recognizing when an individual violates a social or moral norm, such as committing a \emph{faux pas} or acting inappropriately in a given context.
Successful reasoning in this domain requires integrating knowledge of cultural or situational norms with inferences about others’ intentions and emotions. \citet{baron1999recognition} demonstrated that the ability to detect \emph{faux pas} continues to develop through middle childhood, with significant improvements between ages 7, 9, and 11 suggesting that understanding social norm violations requires the integration of both social-norm knowledge and mental-state attribution, a more advanced capacity than basic false-belief reasoning. \cite{strachan2024testing} found that GPT-4 performs comparably to humans across several Theory of Mind tasks, though it shows variability across aspects of social reasoning and particular difficulty with Faux Pas detection. Detecting Social Norm Violations in this study requires the visual input to confirm the exact nature of the agent's conduct which serves as the evidence that must be evaluated against the cultural expectations provided in the scenario text.

\item \textbf{Task $t_5$: Perspective Coordination (PC).}  
This task requires integrating multiple agents’ viewpoints or coordinating what different agents know, see, or believe about a situation. Developmental research shows that children’s ability to coordinate conflicting visual or informational perspectives improves markedly between ages 6–8 \citep{frick2014picturing}. In computational settings, recent works\citet{wilf2024think} demonstrate that explicitly modeling perspective-taking boosts LLM performance on ToM tasks, suggesting that perspective coordination remains a key limitation in the reasoning abilities of current large language models. In this benchmark, Perspective Coordination relies on the visual scene to define distinct physical lines of sight or occlusions, which must be cross-referenced with the informational states described in the text to determine what specific agents can or cannot perceive.

\item \textbf{Task $t_6$: Multi-agent Reasoning (MAR).}
This task involves tracking and reasoning about the mental states of multiple interacting agents simultaneously, including how each agent’s beliefs, intentions, and knowledge may differ from those of others. 
Such scenarios require recursive coordination of nested mental representations across multiple agents, constituting the highest level of cognitive demand within Theory-of-Mind reasoning.
\citet{kim2023fantom} - evaluate over 10,000 multi-party interaction questions and report a substantial human–model performance gap: GPT-4 achieves 26.6\% accuracy compared with 87.5\% for human participants. 
These findings indicate that multi-agent reasoning remains the most challenging and least mastered dimension of ToM for current large language models. For Multi-agent Reasoning, the visual scene is essential to establish the spatial configuration and distinct physical vantage points of multiple characters which serve as the grounding anchor for tracking the recursive mental representations described in the text.
 
\end{enumerate}

\section{Dataset Construction Details}
\label{app:dataset_construction}
\paragraph{Prompt construction.}
For each $(t,c)\in\mathcal{V}$, we define a structured prompt $\mathcal{P}_Q(t,c,\mathcal{D}_i)$ that instructs GPT-4.1 to generate a question testing task $t$ at complexity level $c$. The prompt includes the narrative $d_i$ and the associated cue sets $(E_i,T_i,C_i)$, together with explicit instructions specifying the intended reasoning depth and cognitive focus. The model is directed to generate a single correct answer and three distractors that reflect typical reasoning errors for the target task (for example, mistaking what someone believes for what is actually true in false-belief tasks, or misunderstanding a speaker’s intended meaning in non-literal communication). Options are required to be similar in length and syntactic form, and each question must rely on visual or contextual inference rather than surface-level cues. The complexity level $c$ determines the degree of inference required - from direct perception at Level 1 to recursive multi-agent reasoning at Level 4 - while the task $t$ specifies the mental-state construct being tested. The prompt can be found in Appendix~\ref{app:mcq_prompt}.

\paragraph{Validation and filtering.}
Each generated question underwent a two-stage validation process. 
An automated parser first verified structural integrity - ensuring that each entry contained a single question stem, four labeled options (\texttt{A–D}), one correct answer, and an explanation field - while also checking for formatting consistency and duplicate or empty options. 
Subsequently, two annotators from cultural studies background reviewed all items for clarity and relevance. Crucially, to prevent superficial interpretation, both annotators completed a 40-hour preparatory module prior to validation, covering the cultural backgrounds, region-specific social norms, and contextual cues represented in the selected images, along with calibration exercises on pilot items. The annotation process was independently conducted by the two annotators and subsequently supervised by two PhD researchers, who acted solely as adjudicators for ambiguous cases and ensured that retained items strictly aligned with the intended task–complexity pair $(t, c)$.

\paragraph{Answer randomization.}
\label{app:ans_randomization}
To reduce position bias, the four answer options for each question were randomly permuted using a uniform random permutation $\pi$ over the option indices $\{1,2,3,4\}$. 
Let $O_{i,k}=\{o_{i,1},o_{i,2},o_{i,3},o_{i,4}\}$ denote the original option set and $a_{i,k}$ the correct answer index. 
After randomization, the permuted options and updated answer label are given by
\[
\begin{aligned}
O_{i,k}' &= \{\pi(o_{i,1}),\, \pi(o_{i,2}),\, \pi(o_{i,3}),\, \pi(o_{i,4})\}, \\
a_{i,k}' &= \pi(a_{i,k}).
\end{aligned}
\]
We used a seed value $42$ to ensure reproducibility of option ordering across all generated questions.

\paragraph{Generated Questions.}
Across all $|\mathcal{I}|=394$ images, the procedure produced $N_{\text{total}} = 7092$ questions.  
After automated validation and human review, $N_{\text{valid}} = 5{,}095$ questions were retained, averaging $\bar{q}=12.93$ valid questions per image. We report the dataset statistics in Figure \ref{fig:dataset}.


\paragraph{Human Evaluation.} 
\label{app:human_eval}
To assess the quality and reliability of the constructed dataset, we conducted a rigorous human evaluation on a stratified random sample of approximately 10\% of the total questions ($N=533$). Two independent raters evaluated each item across five key dimensions: answer validity, complexity level correctness, cultural appropriateness, distractor quality, and the necessity of the image for answering the question. Inter-rater reliability was measured using Cohen's Kappa ($\kappa$) \citep{cohen1960coefficient} and agreement percentage, revealing substantial consistency across all metrics. Cultural appropriateness achieved the highest reliability ($\kappa=0.86$, $99.8\%$ agreement), indicating strong consensus on the dataset's cultural fidelity. Similarly, complexity level correctness showed robust alignment ($\kappa=0.67$, $99.8\%$ agreement), validating the distinctiveness of the assigned reasoning levels. The structural quality of the items was also confirmed, with answer validity ($\kappa=0.62$, $92.1\%$ agreement) and distractor quality ($\kappa=0.61$, $89.8\%$ agreement) demonstrating that questions possessed unambiguous correct answers and plausible distractors. Finally, the image requirement dimension ($\kappa=0.62$, $76.7\%$ agreement) verified that the questions effectively test visual reasoning rather than relying solely on textual priors. Note that the divergence between high percent agreement and moderate Kappa scores in these dimensions is an artifact of the prevalence paradox, caused by the high density of positive labels in the dataset which penalizes the Kappa metric.

\section{Supplementary Results}

\subsection{Human Baseline}
\label{app:human_baseline}
 We evaluated human accuracy on a representative subset of the data. We employed a stratified sampling procedure to ensure a balanced evaluation across different question types. Questions were stratified based on Task and Complexity levels. To prevent oversampling underrepresented categories, we enforced an equal number of questions per (Task, Complexity) group, capped by the size of the smallest available group. While the nominal target was 10\% of the dataset (509 questions), the strict balancing constraints resulted in a final sample size of 481 questions.  Human evaluators demonstrated exceptionally high performance and consistency across the sampled dataset. When applying a majority vote aggregation across three raters to resolve individual discrepancies, the human baseline accuracy reached 98.34\% accuracy. We also analyzed the performance of individual raters without aggregation to assess consistency. The raters demonstrated exceptional uniformity, achieving a mean Accuracy of 98.20\% ($\sigma = 0.0063$) and a mean Macro F1 score of 98.21\% ($\sigma = 0.0064$). The marginal difference between the majority vote accuracy ($98.34\%$) and the mean individual accuracy ($98.20\%$) suggests that individual human errors were rare and uncorrelated.

\subsection{Impact of Token Budget on Zero-Shot Chain-of-Thought}
\label{app:zero_shot_cot_token_budget}

We carefully investigated the significant performance drop observed in InternVL 3.5 and Qwen 2.5, when employing zero-shot Chain-of-Thought (CoT) prompting under 256 token budget. We identified that this degradation is primarily caused by insufficient maximum token length during generation. Zero-shot CoT prompts induce substantially longer intermediate reasoning traces, and under a 256-token limit, the models frequently fail to complete the reasoning process or produce a final answer, leading to truncated or invalid outputs. To confirm this hypothesis, we progressively increase the token budget during inference. As shown in Table \ref{tab:vl_token_ablation_global_bold}, increasing the limit to 512 tokens results in a sharp recovery: InternVL 3.5 improves to 89.3\% accuracy on CulturalVQA and 91.3\% on CVQA, while Qwen 2.5 reaches 93.5\% and 92.1\% respectively. Further increasing the limit to 1024 tokens yields additional gains, with InternVL 3.5 achieving 94.1\% on CulturalVQA and Qwen 2.5 maintaining strong performance above 93\% across datasets. These results indicate that the observed performance drop is not due to a failure of reasoning capability, but rather an artifact of token-budget mismatch between prompt style and inference constraints. However, the gain over zero shot prompting is still small compared to older models.

\begin{table*}[h]
\centering
\small
\caption{Zero-shot CoT performance of InternVL and Qwen2.5-VL across data splits under different token limits. Bold values indicate the best score per data split and metric across all settings.}
\label{tab:vl_token_ablation_global_bold}
\begin{tabular}{llcccccc}
\toprule
\multirow{2}{*}{Model} & \multirow{2}{*}{Data Split} &
\multicolumn{2}{c}{256 tokens} &
\multicolumn{2}{c}{512 tokens} &
\multicolumn{2}{c}{1024 tokens} \\
\cmidrule(lr){3-4}\cmidrule(lr){5-6}\cmidrule(lr){7-8}
& & Acc. & Macro F1 & Acc. & F1 & Acc. & F1 \\
\midrule
\multirow{4}{*}{InternVL}
& CulturalVQA   & 0.303 & 0.456 & 0.893 & 0.915 & \textbf{0.941} & \textbf{0.941} \\
& CVQA          & 0.370 & 0.531 & 0.913 & 0.926 & \textbf{0.937} & \textbf{0.937} \\
& FindingEmo   & 0.306 & 0.456 & 0.882 & 0.893 & \textbf{0.905} & \textbf{0.905} \\
\cmidrule(lr){2-8}
& Avg. & 0.319 & 0.473 & 0.891 & 0.905 & \textbf{0.920} & \textbf{0.920} \\
\midrule
\multirow{4}{*}{Qwen2.5-VL}
& CulturalVQA   & 0.389 & 0.549 & \textbf{0.935} & \textbf{0.935} & 0.932 & 0.931 \\
& CVQA          & 0.410 & 0.569 & \textbf{0.921} & \textbf{0.921} & 0.920 & 0.920 \\
& FindingEmo   & 0.386 & 0.542 & \textbf{0.894} & \textbf{0.895} & 0.893 & 0.894 \\
\cmidrule(lr){2-8}
& Avg. & 0.392 & 0.550 & \textbf{0.909} & \textbf{0.910} & 0.907 & 0.908 \\
\bottomrule
\end{tabular}
\end{table*}



    

\begin{figure*}[htbp]
    \centering
    
    \begin{subfigure}{0.49\textwidth}
        \centering
        \includegraphics[width=\textwidth]{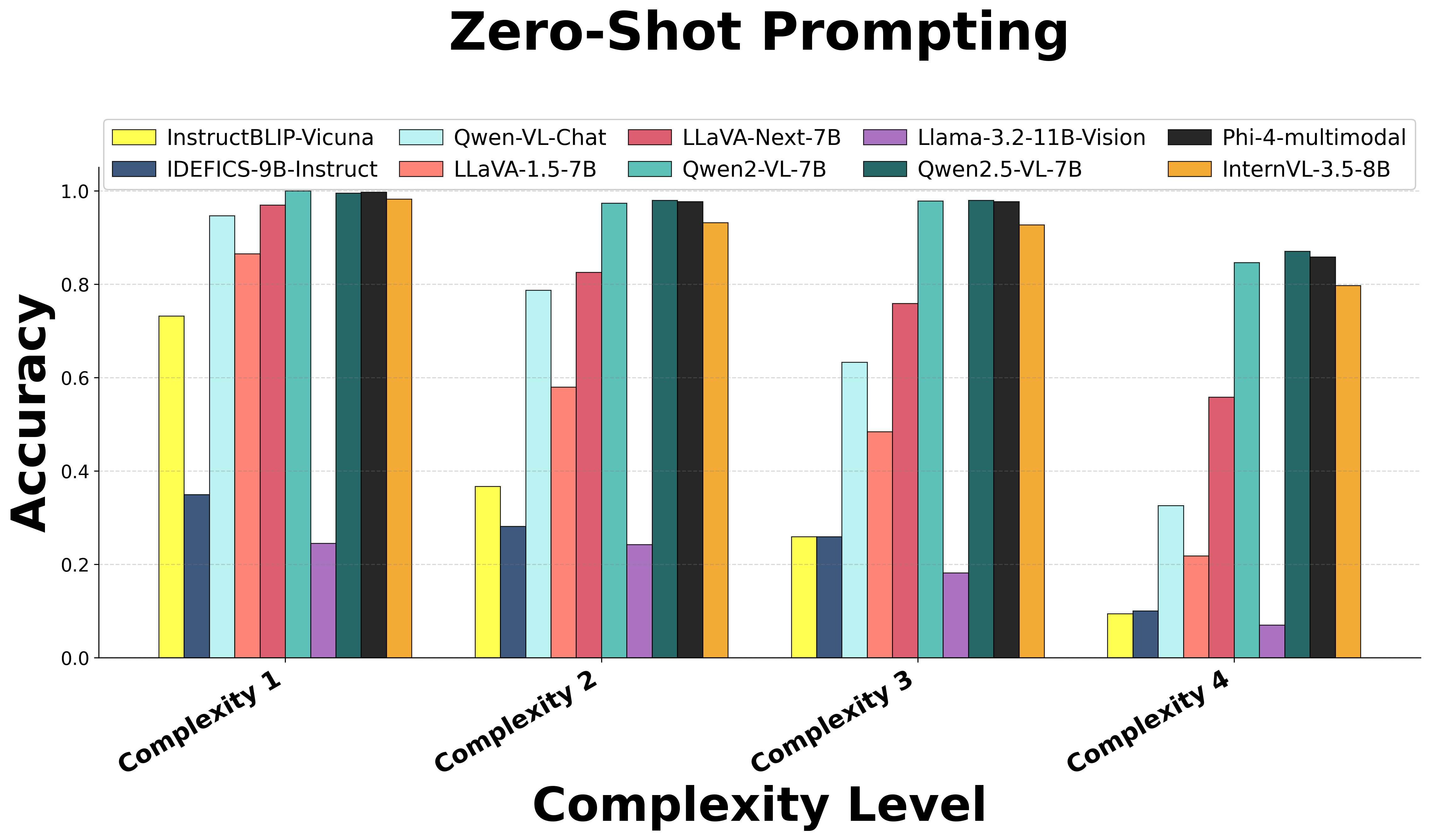}
        \caption{Zero-shot prompting across all complexity levels}
        \label{fig:complexity_zero_shot}
    \end{subfigure}
    \hfill
    \begin{subfigure}{0.49\textwidth}
        \centering
        \includegraphics[width=\textwidth]{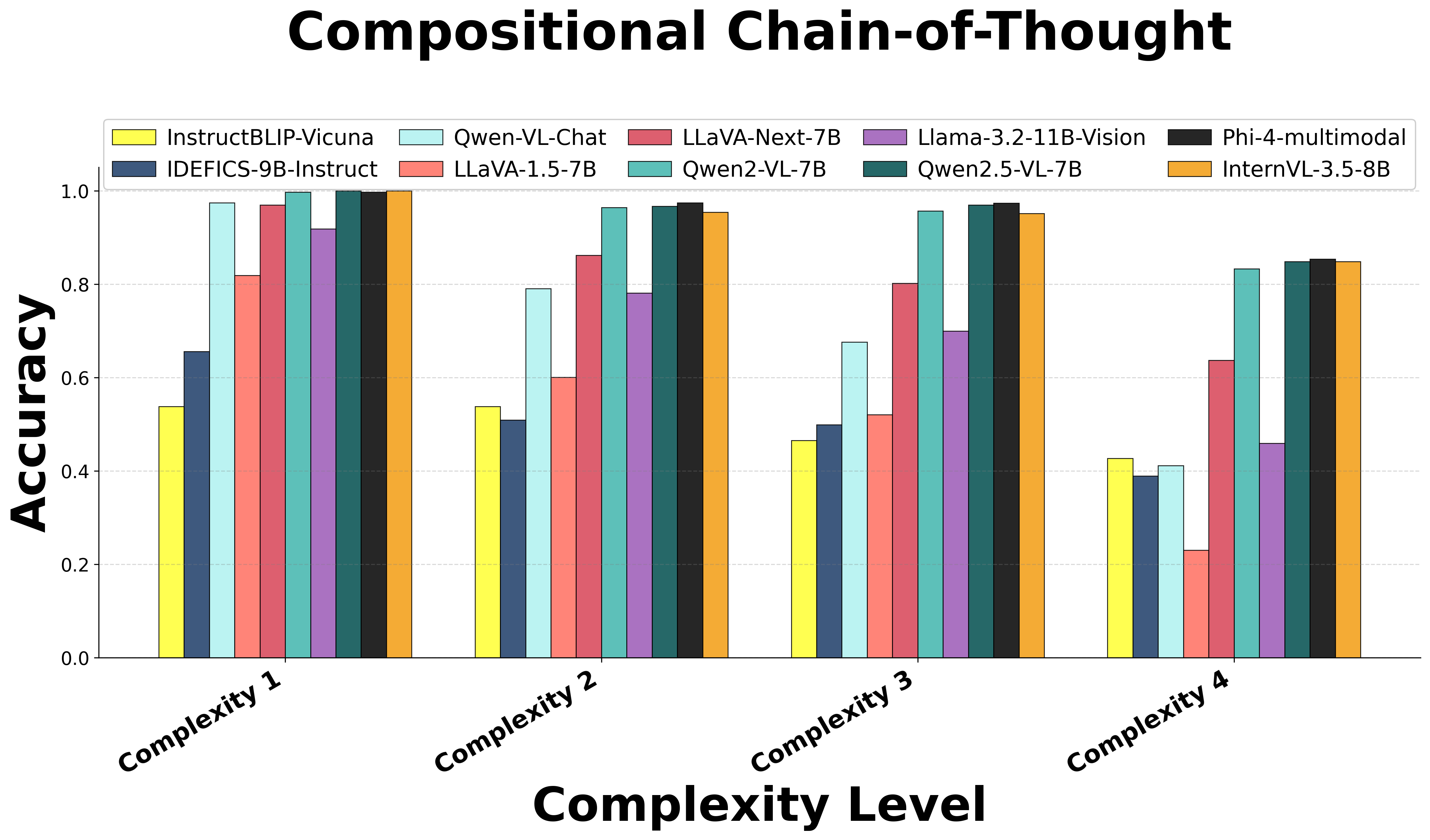}
        \caption{Compositional CoT across all complexity levels}
        \label{fig:complexity_comp_cot}
    \end{subfigure}
    
    \caption{Model performance across four complexity levels of Theory of Mind reasoning.}
    \label{fig:complexity_wise}
\end{figure*}

\subsection{Effect of Option Order Permutation.} Table \ref{tab:qwen25vl_tom_zero_shot} evaluates the robustness of Qwen-2.5-VL (Zero-Shot) to permutations in multiple-choice option order. Permuting the option order results in minimal performance variation, indicating that the model is largely robust to option-order bias. On CulturalVQA, shuffling options slightly improves performance, yielding a marginal gain of +0.35\% in both accuracy and F1. For CVQA, we observe a small degradation (-0.55\% accuracy, -1.31\% F1), suggesting mild sensitivity but no substantial collapse. On FindingEmo, performance remains effectively unchanged, with near-zero deltas across metrics. These results indicate that while minor fluctuations exist, option order permutation does not meaningfully degrade performance for Qwen-2.5-VL. The limited magnitude of performance change suggests that the model relies primarily on semantic reasoning rather than positional heuristics when selecting answers - which is expected given that we randomized model generated answers while constructing the dataset.

\begin{table*}[t]
\centering
\small

\caption{Performance of Qwen2.5 VL Zero-Shot on cross-cultural Theory-of-Mind data splits. The $\Delta$ column indicates performance difference compared to the original (non-shuffled) option order.}
\label{tab:qwen25vl_tom_zero_shot}

\begin{tabular}{l l cc cc}
\toprule
\multirow{2}{*}{Image source} & \multirow{2}{*}{Option Order} 
& \multicolumn{2}{c}{Accuracy} 
& \multicolumn{2}{c}{F1} \\
& & Value & $\Delta$ & Value & $\Delta$ \\
\midrule

\multirow{2}{*}{\textbf{CulturalVQA}} 
&  \textit{Shuffled} & 0.9573 & +0.0035 & 0.9572 & +0.0035 \\
& Original    & 0.9538 & ---     & 0.9537 & ---     \\
\midrule

\multirow{2}{*}{\textbf{CVQA}}
& \textit{Shuffled} & 0.9416 & -0.0055 & 0.9339 & -0.0131 \\
& Original    & 0.9471 & ---     & 0.9470 & ---     \\
\midrule

\multirow{2}{*}{\textbf{FindingEmo}}
& \textit{Shuffled} & 0.9294 & 0.0000 & 0.9291 & -0.0002 \\
& Original    & 0.9294 & ---    & 0.9293 & ---     \\
\bottomrule
\end{tabular}

\end{table*}





\section{Model Access URLs}
Table~\ref{tab:hug_model_urls} lists the official repository URLs for all models evaluated in this study.
\label{app:model_urls}

\begin{table*}[ht]
\centering
\footnotesize
\setlength{\tabcolsep}{6pt}
\renewcommand{\arraystretch}{1.1}
\begin{tabular}{ll}
\toprule
\textbf{Model} & \textbf{URL} \\
\midrule
InternVL-3.5-8B &
\href{https://huggingface.co/OpenGVLab/InternVL3_5-8B-Instruct}{InternVL-3.5-8B} \\

LLaVA-1.5-7B &
\href{https://huggingface.co/llava-hf/llava-1.5-7b-hf}{LLaVA-1.5-7B} \\

LLaVA-Next-7B &
\href{https://huggingface.co/llava-hf/llava-v1.6-mistral-7b-hf}{LLaVA-Next-7B} \\

Qwen-VL Chat &
\href{https://huggingface.co/Qwen/Qwen-VL-Chat}{Qwen-VL Chat} \\

Qwen2-VL-7B &
\href{https://huggingface.co/Qwen/Qwen2-VL-7B-Instruct}{Qwen2-VL-7B} \\

Qwen2.5-VL-7B &
\href{https://huggingface.co/Qwen/Qwen2.5-VL-7B-Instruct}{Qwen2.5-VL-7B} \\

Phi-4-6B &
\href{https://huggingface.co/microsoft/Phi-4-multimodal-instruct}{Phi-4-6B} \\

InstructBLIP-Vicuna &
\href{https://huggingface.co/Salesforce/instructblip-vicuna-7b}{InstructBLIP-Vicuna} \\

IDEFICS-9B &
\href{https://huggingface.co/HuggingFaceM4/idefics-9b-instruct}{IDEFICS-9B} \\

Llama-11B &
\href{https://huggingface.co/meta-llama/Llama-3.2-11B-Vision-Instruct}{Llama-11B} \\
\bottomrule
\end{tabular}
\caption{Model access links used in this work.}
\label{tab:hug_model_urls}
\end{table*}

\section{Supplementary Analyses: Lexical Shortcuts and Positivity Bias}
\label{app:positivity_bias}

This appendix reports supplementary analyses supporting the main results on text-only performance and social desirability tendencies. We first evaluate whether surface-level lexical cues alone can solve the task. We then quantify sentiment skew in the dataset itself, followed by model-level diagnostics measuring whether predictions preferentially align with semantically positive answer options in the absence of visual input.

\subsection{Text-only Baseline: TF--IDF with Logistic Regression}
\label{app:tfidf_lr}

\paragraph{Setup.}
We evaluate a purely textual baseline using TF--IDF features with multinomial logistic regression (LR) for the 4-way multiple-choice task.
Each example is represented as $(x_i, y_i)$, where $y_i \in \{1,2,3,4\}$ denotes the index of the correct answer option.
We consider two input representations:
\begin{align}
x_i^{(\text{Q+O})} &= \texttt{concat}(q_i, o_{i,1}, o_{i,2}, o_{i,3}, o_{i,4}), \\
x_i^{(\text{O})} &= \texttt{concat}(o_{i,1}, o_{i,2}, o_{i,3}, o_{i,4}),
\end{align}
where $q_i$ is the question stem and $o_{i,j}$ denotes the text of option $j$.
We perform 5-fold cross-validation and report mean accuracy and standard deviation across folds.
Chance performance is $0.25$.

\begin{table}[t]
\centering
\small
\caption{TF--IDF + Logistic Regression text-only baseline (4-way classification). Reported as mean $\pm$ standard deviation across 4 folds.}
\label{tab:tfidf_lr}
\begin{tabular}{lcc}
\toprule
\textbf{Split} & \textbf{Question + Options} & \textbf{Options-only} \\
\midrule
CulturalVQA & $0.233 \pm 0.020$ & $0.244 \pm 0.016$ \\
CVQA        & $0.257 \pm 0.020$ & $0.261 \pm 0.023$ \\
FindingEmo  & $0.252 \pm 0.014$ & $0.261 \pm 0.020$ \\
\bottomrule
\end{tabular}
\end{table}

\subsection{Dataset-level Positivity Skew in Gold Answers}
\label{app:dataset_pos_bias}

\paragraph{Sentiment scoring.}
We use a pretrained sentiment classifier from the \texttt{transformers} library (\href{https://huggingface.co/cardiffnlp/twitter-roberta-base-sentiment-latest}
{\texttt{twitter-roberta-sentiment}}) for sentiment scoring. Each answer option $o$ is assigned a positivity score using:
\begin{equation}
\text{pos}(o) = P(\text{positive} \mid o).
\end{equation}
We also define a signed sentiment margin:
\begin{equation}
\text{margin}(o) = P(\text{positive} \mid o) - P(\text{negative} \mid o).
\end{equation}

\paragraph{Dataset diagnostics.}
For each example $i$ with option set $O_i = \{o_{i,1}, \dots, o_{i,4}\}$ and gold label $y_i$, we compute three dataset-level diagnostics that ignore model predictions:

\paragraph{Top-positivity accuracy (margin-based).}
\begin{equation}
\begin{aligned}
\mathrm{Acc}_{\text{margin}}
&=
\frac{1}{N}\sum_{i=1}^{N}
\mathbb{I}\!\Big[
y_i =
\arg\max_{j \in \{1,\dots,4\}} \\
&\qquad
\text{margin}(o_{i,j})
\Big]
\end{aligned}
\end{equation}
with a random-choice baseline of $0.25$.

\paragraph{Top-positivity accuracy (positivity-only).}
\begin{equation}
\mathrm{Acc}_{\text{pos}} =
\frac{1}{N}\sum_{i=1}^{N}
\mathbb{I}\!\left[
y_i = \arg\max_{j \in \{1,\dots,4\}} \text{pos}(o_{i,j})
\right],
\end{equation}
also with a baseline of $0.25$.

\paragraph{Gold-option positivity dominance.}
Let $\mathcal{D}_i = \{j \neq y_i\}$ denote the three distractor indices. We define:
\begin{equation}
\mathrm{Dom}_{\text{pos}} =
\frac{1}{N}\sum_{i=1}^{N}
\mathbb{I}\!\left[
\text{pos}(o_{i,y_i}) >
\frac{1}{3}\sum_{j \in \mathcal{D}_i}\text{pos}(o_{i,j})
\right],
\end{equation}
with a baseline of $0.50$ under random dominance.

\begin{table}[t]
\centering
\small
\caption{Dataset-level positivity diagnostics computed from gold answers using a sentiment classifier.}
\label{tab:dataset_pos_metrics}
\begin{tabular}{lccc}
\toprule
\textbf{Split} & $\mathrm{Acc}_{\text{margin}}$ & $\mathrm{Acc}_{\text{pos}}$ & $\mathrm{Dom}_{\text{pos}}$ \\
\midrule
CulturalVQA  & $0.5641$ & $0.5615$ & $0.6650$ \\
CVQA         & $0.5482$ & $0.5436$ & $0.6614$ \\
FindingEmo  & $0.4556$ & $0.4542$ & $0.5409$ \\
\bottomrule
\end{tabular}
\end{table}

\subsection{Model-level Positivity Alignment in the No-image Setting}
\label{app:model_pos_bias}

\paragraph{Models, setting, and outputs.}
We analyze three vision--language models: \texttt{Qwen-VL-Chat}, \texttt{Qwen2-VL-7B}, and \texttt{Qwen2.5-VL-7B}.
All results are obtained in the text-only (no-image) zero-shot setting.
Each example $i$ is a 4-way multiple-choice item with options $\{o_{i,0},o_{i,1},o_{i,2},o_{i,3}\}$ and gold label $y_i \in \{0,1,2,3\}$ (mapped from \{A,B,C,D\}).

\paragraph{Option positivity scores.}
Each option $o_{i,j}$ is assigned a positivity score
\(
\mathrm{pos}(o_{i,j}) \in [0,1]
\),
defined as the sentiment model probability $P(\text{positive}\mid o_{i,j})$.
Let
\begin{equation}
j_i^{\ast} = \arg\max_{j \in \{0,1,2,3\}} \mathrm{pos}(o_{i,j})
\end{equation}
denote the index of the most-positive option.

\paragraph{Resolvable set.}
Let $\hat{y}_i \in \{0,1,2,3\}$ denote the predicted option index obtained by an explicit option letter.
Let $\mathcal{R}$ denote the set of items for which $\hat{y}_i$ is resolvable.

\paragraph{Most-positive option selection rate.}
We compute:
\begin{equation}
\mathrm{Sel}_{\text{top-pos}} =
\frac{1}{|\mathcal{R}|}
\sum_{i \in \mathcal{R}}
\mathbb{I}[\hat{y}_i = j_i^{\ast}],
\end{equation}
with a random-choice baseline of $0.25$.

\paragraph{Excess positivity on incorrect predictions.}
Let $\mathcal{W} = \{i \in \mathcal{R} : \hat{y}_i \neq y_i\}$.
For each $i \in \mathcal{W}$, define:
\begin{align}
\mathrm{Obs}_i &= \mathbb{I}[\mathrm{pos}(o_{i,\hat{y}_i}) > \mathrm{pos}(o_{i,y_i})], \\
\mathrm{CF}_i  &= \frac{1}{3}\sum_{j \in \{0,1,2,3\}\setminus \{y_i\}}
\mathbb{I}[\mathrm{pos}(o_{i,j}) > \mathrm{pos}(o_{i,y_i})].
\end{align}
The excess positivity rate is:
\begin{equation}
\mathrm{Excess}_{\text{pos}} =
\frac{1}{|\mathcal{W}|}\sum_{i \in \mathcal{W}} (\mathrm{Obs}_i - \mathrm{CF}_i),
\end{equation}
with a no-drift baseline of $0$.

\paragraph{Excess selection of the most-positive distractor under conflict.}
Define the set of conflict items:
\begin{equation}
\mathcal{C} = \{i \in \mathcal{R} : j_i^{\ast} \neq y_i\},
\end{equation}
and restrict to items that are both conflict and incorrect:
$\mathcal{CW} = \mathcal{C} \cap \mathcal{W}$.
We compute:
\begin{equation}
\mathrm{Excess}_{\text{conflict}} =
\frac{1}{|\mathcal{CW}|}
\sum_{i \in \mathcal{CW}}
\left(
\mathbb{I}[\hat{y}_i = j_i^{\ast}] - \frac{1}{3}
\right),
\end{equation}
where $\frac{1}{3}$ is the uniform baseline among the three distractors.

\paragraph{Uncertainty estimates.}
For $\mathrm{Excess}_{\text{pos}}$ and $\mathrm{Excess}_{\text{conflict}}$, we report 95\% bootstrap confidence intervals computed over the corresponding per-item excess values using percentile bootstrap with 2,000 resamples.

\begin{table*}[t]
\centering
\small
\caption{
Selection of positive-sentiment answer options in the text-only setting.
The first column reports how often a model selects the most positive option overall
(random baseline: 25\%).
The second reports excess positivity on incorrect predictions, measuring whether
wrong answers are more positive than the gold answer beyond chance (random baseline: 0).
The third further restricts to incorrect cases where the correct answer is not the
most positive option, measuring excess selection of the most positive distractor
(random baseline: 0).
Error bars in brackets indicate 95\% bootstrap confidence intervals.
}
\label{tab:positivity_bias_simple}
\begin{tabular}{llccc}
\toprule
\textbf{Image source} & \textbf{Model} &
\textbf{Top-positive sel.} &
\textbf{Excess positivity (wrong)} &
\textbf{Excess positivity (conflict)} \\
\midrule
CulturalVQA
& Qwen-VL-Chat
& 0.484
& 0.038 [0.008, 0.066]
& 0.097 [0.032, 0.162] \\
& Qwen2-VL-7B
& 0.568
& 0.114 [0.035, 0.193]
& 0.204 [0.074, 0.333] \\
& Qwen2.5-VL-7B
& 0.556
& 0.111 [0.037, 0.196]
& 0.222 [0.056, 0.389] \\
\midrule
CVQA
& Qwen-VL-Chat
& 0.456
& 0.010 [$-$0.018, 0.039]
& 0.022 [$-$0.048, 0.091] \\
& Qwen2-VL-7B
& 0.536
& 0.079 [0.000, 0.153]
& 0.194 [0.028, 0.361] \\
& Qwen2.5-VL-7B
& 0.531
& 0.005 [$-$0.067, 0.077]
& 0.045 [$-$0.117, 0.207] \\
\midrule
FindingEmo
& Qwen-VL-Chat
& 0.386
& 0.009 [$-$0.007, 0.026]
& 0.030 [$-$0.006, 0.067] \\
& Qwen2-VL-7B
& 0.466
& 0.076 [0.038, 0.114]
& 0.150 [0.081, 0.218] \\
& Qwen2.5-VL-7B
& 0.448
& 0.046 [0.010, 0.081]
& 0.090 [0.017, 0.164] \\
\bottomrule
\end{tabular}
\end{table*}

\paragraph{Analysis.} From \ref{tab:positivity_bias_simple}, across all splits, models exhibit a systematic tendency to prefer positive-sentiment answer options even in the absence of visual input. Overall, the probability of selecting the most positive option substantially exceeds the random baseline of 25\%, ranging from 38.6\% to 56.8\% across models and datasets - corresponding to an absolute increase of approximately 14–32 percentage points over chance. This tendency persists when restricting to incorrect predictions: excess positivity on errors is positive for most model–dataset pairs, with statistically supported effects in several cases (e.g., +11.4\% for Qwen2-VL-7B and +11.1\% for Qwen2.5-VL-7B on CulturalVQA), indicating that when models err, they are more likely than chance to select an option that is more positive than the correct answer. The strongest evidence appears under conflict conditions, where the correct answer is not the most positive option: here, models select the most positive distractor 5–22 percentage points more often than expected under uniform random choice among distractors. For example, on CulturalVQA, Qwen2.5-VL-7B chooses the most positive distractor approximately 55.6\% of the time in conflict-and-wrong cases, compared to a random baseline of 33.3\%. While the magnitude of this effect varies across datasets and is weaker or statistically inconclusive for some settings (notably CVQA and FindingEmo with Qwen-VL-Chat), the overall pattern indicates a consistent, though partial, positivity-aligned heuristic that increases the likelihood of selecting socially favorable answers beyond chance, without fully accounting for the high accuracies observed for larger models.

\section{Qualitative Analysis of Predictions without Visual Context}
\label{app:qual_text_only}

We present representative qualitative examples illustrating how vision--language
models produce confident yet image-dependent predictions when evaluated without the visual context. In both cases, models converge on a single answer despite the
absence of visual input, whereas human annotators correctly identify the questions
as unanswerable.

\begin{vlqa}
\scriptsize
\textbf{Question:}
Based on the group’s visible expressions and gestures, what is the most likely feeling shared among the children?

\vspace{2pt}
\textbf{Options:}
\begin{enumerate}[label=\Alph*), leftmargin=2.35em, itemsep=1pt, topsep=1pt]
  \item They are expressing disappointment in the activity.
  \item They are showing nervousness about being outside.
  \item They are feeling confused about their surroundings.
  \item They are experiencing happiness together.
\end{enumerate}

\tcblower
\textbf{Predictions}\vspace{1pt}

\predcellrow{Qwen-VL}{D}
\predcellrow{Qwen-2}{D}
\predcellrow{Qwen-2.5}{D}
\predcellhuman{Human}{Not answerable without image}
\end{vlqa}

\begin{vlqa}
\scriptsize
\textbf{Question:}
Based on the boy's posture and expression while he sits in the basin, what is the most likely feeling he is experiencing in this situation?

\vspace{2pt}
\textbf{Options:}
\begin{enumerate}[label=\Alph*), leftmargin=2.35em, itemsep=1pt, topsep=1pt]
  \item He is feeling worried about being outside in the basin.
  \item He is feeling comfortable and at ease in his environment.
  \item He is feeling bored and uninterested in the activity.
  \item He is feeling upset about being watched by someone nearby.
\end{enumerate}

\tcblower
\textbf{Predictions}\vspace{1pt}

\predcellrow{Qwen-VL}{B}
\predcellrow{Qwen-2}{B}
\predcellrow{Qwen-2.5}{B}
\predcellhuman{Human}{Not answerable without image}
\end{vlqa}

\section{Prompts}
\label{app:prompts}

\subsection{Scene Description and Cue Extraction Prompt}
\label{app:scene_prompt}
\begin{tcolorbox}[
    enhanced,
    colback=gray!3,
    colframe=black!50,
    boxrule=0.6pt,
    arc=3pt,
    left=6pt,
    right=6pt,
    top=4pt,
    bottom=4pt,
    fonttitle=\bfseries,
    title={Prompt Template for Generating ToM-Focused Scene Descriptions},
    breakable
]
\begin{PromptVerb}
[Image]

You are an expert in Theory of Mind (ToM) reasoning and cultural analysis.
Analyze the provided image and the brief description in the metadata, and
infer a concise yet comprehensive ToM-focused description of the scene.
Emphasize emotional states, interpersonal interactions, and situational
dynamics relevant to Theory of Mind and representative culture. Use the
metadata as additional context to enrich your description.

Additionally, extract specific cues under the following categories:
1. EmotionalCues: Key phrases that capture the emotional states visible
   or implied in the scene.
2. ToMCues: Key phrases that capture Theory of Mind-related elements such
   as intentions, beliefs, or interpersonal relationships. Emphasize
   culture-specific ToM where applicable.
3. CulturalCues: Key phrases that highlight culture type, cultural symbols,
   norms, or practices depicted in the image.

Ensure the output is formatted as a structured JSON object.

Input:
{
    "Metadata": "[Insert the description from metadata]"
}

Output:
{
    "ToMFocusedDescription": "[Insert inferred ToM-focused description
        based on the input]",
    "EmotionalCues": ["Cue 1", "Cue 2", ...],
    "ToMCues": ["Cue 1", "Cue 2", ...],
    "CulturalCues": ["Cue 1", "Cue 2", ...]
}
\end{PromptVerb}
\end{tcolorbox}

\subsection{Theory of Mind Multiple-Choice Question Generation Prompt}
\label{app:mcq_prompt}
\begin{tcolorbox}[
    enhanced,
    colback=gray!3,
    colframe=black!50,
    boxrule=0.6pt,
    arc=3pt,
    left=6pt,
    right=6pt,
    top=4pt,
    bottom=4pt,
    fonttitle=\bfseries,
    title={Prompt Template for Generating ToM-Based Multiple-Choice Questions},
    breakable
]
\begin{PromptVerb}
[Image]


You are an expert in multi-cultural Theory of Mind reasoning.

TASK:
Write a multiple-choice question that evaluates the Theory of Mind task "{task}" 
at complexity level {complexity}.

COMPLEXITY LEVEL:
Level {complexity}: {complexity_description}


INSTRUCTIONS:
1. Return exactly one question for complexity level {complexity} focusing on 
   the task "{task}" in JSON format.
2. Include 4 answer choices per question (labeled A through D) with one 
   objectively correct response.
3. Make the distractors plausible and commonly confusable with the correct 
   answer  -  prefer near-misses (e.g., flipped intentions or misread cultural 
   norms). Ensure they’re mutually exclusive, similar in length and tone to 
   the correct option, and avoid generic negations, tautologies, or irrelevant 
   details.
4. Mark the correct answer (A, B, C, or D).
5. Provide a brief explanation of how the question measures the specific ToM task.
6. Do NOT describe the image, the emotion or activities of the characters, or 
   restate its caption in the question stem. Instead, refer to the main 
   character(s) in neutral or role-based terms so that the test-taker, who 
   only sees the image and the question, must rely on interpreting the visual 
   scene itself rather than being handed details from the image description.
7. The question MUST not contain any direct hints or giveaways to the correct answer.
8. Incorporate cultural context appropriately.


REQUIREMENTS:
- Base the question on the provided image, image description, cues, and 
  cultural context.
- Use details from Emotional/ToM/Cultural Cues in the question or explanation.
- Design distractors as near-misses: each wrong option must sound plausible 
  and differ from the correct answer only by a subtle mistake in belief, 
  perspective, or norm interpretation.
- For complexity 3 and 4, optionally anchor the scenario in an ANALOGOUS, 
  less-common cultural setting if supported by Cultural Cues (e.g., map a 
  familiar ritual to an analogous practice noted in the cues); keep the 
  mental-state logic consistent.
- The question should accurately reflect the specified complexity level.
- Maintain consistent formatting in your response.


OUTPUT SCHEMA:
Return the question as structured JSON:

{
  "Question": <question text>,
  "Options": ["A) option1", "B) option2", "C) option3", "D) option4"],
  "CorrectAnswer": <letter A-D>,
  "Task": "{task}",
  "Complexity": {complexity},
  "Explanation": <brief reasoning for how the question tests the Theory of 
                  Mind task and references provided cues>
}


CONTEXT:
Image Description: {{description}}
Emotional Cues: {{emotional_cues}}
Theory of Mind Cues: {{tom_cues}}
Cultural Cues: {{cultural_cues}}
\end{PromptVerb}
\end{tcolorbox}

\subsection{Zero-shot Prompt}
\label{app:zero_shot}
\begin{tcolorbox}[
    enhanced,
    colback=gray!3,
    colframe=black!50,
    boxrule=0.6pt,
    arc=3pt,
    left=6pt,
    right=6pt,
    top=4pt,
    bottom=4pt,
    fonttitle=\bfseries,
    title={Zero-shot Prompt Template},
    breakable
]
\begin{PromptVerb}
[IMAGE]

Question: <text>
Options:
A) ...
B) ...
C) ...
D) ...

Answer: (Write only the option letter, e.g., "A")
\end{PromptVerb}
\end{tcolorbox}

\subsection{Chain-of-Thought (CoT) Prompt}
\label{app:zero_shot_cot}
\begin{tcolorbox}[
    enhanced,
    colback=gray!3,
    colframe=black!50,
    boxrule=0.6pt,
    arc=3pt,
    left=6pt,
    right=6pt,
    top=4pt,
    bottom=4pt,
    fonttitle=\bfseries,
    title={Zero-shot Chain-of-Thought Prompt Template},
    breakable
]
\begin{PromptVerb}
[IMAGE]

Question: <text>
Options:
A) ...
B) ...
C) ...
D) ...

Let's think step by step.
Finally, give your answer in the format: "Answer: <option letter>"
\end{PromptVerb}
\end{tcolorbox}

\subsection{Compositional Chain-of-Thought (CoT) Prompts}
\label{app:compositional_cot}
\begin{tcolorbox}[
    enhanced,
    colback=gray!3,
    colframe=black!50,
    boxrule=0.6pt,
    arc=3pt,
    left=6pt,
    right=6pt,
    top=4pt,
    bottom=4pt,
    fonttitle=\bfseries,
    title={Compositional Chain-of-Thought: Scene Graph Construction},
    breakable
]
\begin{PromptVerb}
[IMAGE]

Task: {question}
Options: A) ..., B) ..., C) ..., D) ...

For the provided image and its associated question, generate a scene graph in JSON
format that includes:
1. Objects relevant to answering the question.
2. Attributes relevant to answering the question.
3. Relationships among the relevant objects.

Scene Graph:
\end{PromptVerb}
\end{tcolorbox}

\begin{tcolorbox}[
    enhanced,
    colback=gray!3,
    colframe=black!50,
    boxrule=0.6pt,
    arc=3pt,
    left=6pt,
    right=6pt,
    top=4pt,
    bottom=4pt,
    fonttitle=\bfseries,
    title={Compositional Chain-of-Thought: Answer Generation},
    breakable
]
\begin{PromptVerb}
[IMAGE]

Scene Graph: ...

Use the image and scene graph as context and answer the following question:

Task: {question}
Options: A) ..., B) ..., C) ..., D) ...

Answer with the option letter from the given choices directly.
Answer:
\end{PromptVerb}
\end{tcolorbox}

\end{document}